%% file: main.tex
\documentclass{article}
\pdfoutput=1

\usepackage{PRIMEarxiv}

\usepackage[utf8]{inputenc} 
\usepackage[T1]{fontenc}    
\usepackage{hyperref}       
\usepackage{url}            
\usepackage{booktabs}       
\usepackage{amsfonts}       
\usepackage{nicefrac}       
\usepackage{microtype}      
\usepackage{lipsum}
\usepackage{fancyhdr}       
\usepackage{graphicx}       
\graphicspath{{media/}}     

\usepackage{balance} 
\usepackage{graphicx}
\usepackage{subcaption}
\usepackage{mathtools}
\usepackage{amsthm}
\usepackage{booktabs}
\usepackage{colortbl}
\usepackage{subcaption}
\usepackage{wrapfig}
\usepackage{booktabs}
\usepackage{multirow}

\usepackage{appendix}
\usepackage{pgfplots}
\pgfplotsset{compat=1.10}
\usepackage{adjustbox}
\usepackage{pdfpages}
\usepackage{appendix}
\usepackage{etoc}
\usepackage{booktabs}

\pgfplotsset{compat=1.17}

\usepackage[disable]{todonotes}
\newcommand{\ava}[1]{\todo[backgroundcolor=teal!20, linecolor=teal!85!black]{\textbf{Ava:} #1}}

\newcommand{\hb}[1]{\todo[backgroundcolor=blue!10, linecolor=blue!50!black]{\textbf{Hendrik:} #1}}
\usepackage[normalem]{ulem}

\usepackage[ruled,vlined,linesnumbered]{algorithm2e}
\usepackage[capitalize]{cleveref}
\usepackage[font=small,labelfont=bf]{caption}
\usepackage{hyperref}       
\usepackage{url}            
\usepackage{booktabs}       
\usepackage{amsfonts}       
\usepackage{nicefrac}       
\usepackage{microtype}      
\usepackage{xcolor}         
\usepackage{natbib}

\newtheorem{prop}{Proposition}

\newtheorem{theorem}{Theorem}
\newtheorem{corollary}{Corollary}[theorem]

\usepackage{tkz-euclide}
\DeclareMathOperator*{\argmax}{arg\,max}

\title{Decision Making in Non-Stationary Environments with Policy-Augmented Search
\thanks{Recommended Citation: Pettet, Ava, Zhang, Yunuo, Luo, Baiting, Wray, Kyle, Baier, Hendrik, Laszka, Aron, Dubey, Abhishek, and Mukhopadhyay, Ayan. ``Decision Making in Non-Stationary Environments with Policy-Augmented Search.'' \textit{International Conference on Autonomous Agents and MultiAgent Systems (AAMAS)}. 2023.}
}

\author{
  Ava Pettet\textsuperscript{*}\\
  Vanderbilt University \\
  Nashville, TN \\
  \And
  Yunuo Zhang\textsuperscript{*} \\
  Vanderbilt University \\
  Nashville, TN \\
   \And
  Baiting Luo \\
  Vanderbilt University \\
  Nashville, TN \\
   \And
  Kyle Wray \\
  Stanford University \\
  Palo Alto, CA \\
   \And
  Hendrik Baier \\
  Eindhoven University of Technology \\
  Eindhoven, Netherlands \\
   \And
  Aron Laszka \\
  Pennsylvania State University \\
  University Park, PA \\
   \And
  Abhishek Dubey\\
  Vanderbilt University \\
  Nashville, TN \\
  \And
  Ayan Mukhopadhyay\\
  Vanderbilt University \\
  Nashville, TN \\\\
  \texttt{\{ava.pettet, yunuo.zhang, baiting.luo, abhishek.dubey, ayan.mukhopadhyay\}@vanderbilt.edu}\\
  \texttt{kyle.hollins.wray@gmail.com}, \texttt{h.baier@tue.nl}, \texttt{aron.laszka@psu.edu}
    \\\\
  \textsuperscript{*}These authors contributed equally to this work.
}

\begin{document}
\maketitle

\input{abstract}



         
\newcommand{\BibTeX}{\rm B\kern-.05em{\sc i\kern-.025em b}\kern-.08em\TeX}


\input{Introduction}

\input{background}
\input{Problem}
\input{approach}
\input{proof}
\input{experiments_ayan}
\input{relatedWork}
\input{conclusion}

\section*{Acknowledgments}
This material is based upon work sponsored by the National Science Foundation (NSF) under Grant CNS-2238815 and by the Defense Advanced Research Projects Agency (DARPA). We also acknowledge Any opinions, findings, and conclusions or recommendations expressed in this material are those of the authors and do not necessarily reflect the views of the NSF, or the DARPA. Results presented in this paper were obtained using the Chameleon testbed supported by the National Science Foundation.

\bibliographystyle{plainnat}
\bibliography{refs}

\newpage
\pagenumbering{arabic}
\input{appendix}

\end{document}

%% file: abstract.tex
\begin{abstract}
    Sequential decision-making under uncertainty is present in many important problems. Two popular approaches for tackling such problems are reinforcement learning and online search (e.g., Monte Carlo tree search). While the former learns a policy by interacting with the environment (typically done before execution), the latter uses a generative model of the environment to sample promising action trajectories at decision time. Decision-making is particularly challenging in non-stationary environments, where the environment in which an agent operates can change over time. Both approaches have shortcomings in such settings---on the one hand, policies learned before execution become stale when the environment changes and relearning takes both time and computational effort. Online search, on the other hand, can return sub-optimal actions when there are limitations on allowed runtime. In this paper, we introduce \textit{Policy-Augmented Monte Carlo tree search} (PA-MCTS), which combines action-value estimates from an out-of-date policy with an online search using an up-to-date model of the environment. We prove theoretical results showing conditions under which PA-MCTS selects the one-step optimal action and also bound the error accrued while following PA-MCTS as a policy. 
    We compare and contrast our approach with AlphaZero, another hybrid planning approach, and Deep Q Learning on several OpenAI Gym environments. Through extensive experiments, we show that under non-stationary settings with limited time constraints, PA-MCTS outperforms these baselines. 
\end{abstract}

%% file: Introduction.tex
\section{Introduction}\label{sec:intro}

Sequential decision-making is present in many important problem domains, such as autonomous driving~\citep{bouton2019safe}, emergency response~\citep{mukhopadhyay2019online}, and medical diagnosis~\citep{ayer2012or}. An open challenge in such settings is non-stationary environments, where the dynamics of the environment can change over time. A decision agent must adapt to these changes to avoid taking sub-optimal actions. 
Two well-known approaches for sequential decision-making are reinforcement learning (RL) and online planning~\citep{kochenderfer2022algorithms}. In RL approaches, an agent learns a policy $\pi$, i.e., a mapping from states to actions, through interacting with the environment. 
The learning can also take place before execution using environmental models. Once a policy is learned, it can be invoked nearly instantaneously at decision time.
Deep RL methods, which use a neural network as a function approximator for the policy, have achieved state-of-the-art performance in many applications~\citep{silver_mastering_2016,kochenderfer2022algorithms}. However, when faced with non-stationary environments, a policy can become stale and result in sub-optimal decisions. Moreover, retraining the policy on the new environment takes time and considerable computational effort, particularly in problems with complex state-action spaces. While RL algorithms have been designed to operate in non-stationary environments~\citep{ortner2020variational,cheung2019non}, there is a delay between when a change is detected and when the learning framework converges to the updated policy. Depending on the problem setting, such delays might be very expensive.

An alternative approach is using algorithms such as Monte Carlo tree search (MCTS) to perform online planning. These approaches perform their computation at decision time using high-fidelity models of the current environment to find promising action trajectories. These models can be updated as soon as environmental changes are detected, and such changes can be immediately incorporated into decision-making (assuming that the generative model used to build the search tree can be updated quickly). MCTS has been proven to converge to optimal actions given enough computation time~\citep{kocsis2006bandit}, but convergence can be slow for domains with large state-action spaces. The slow convergence is a potential issue for problem settings with tight constraints on the time allowed for decision-making; e.g., in emergency response, when an incident occurs, any time used for decision-making increases the overall response time~\citep{pettet2021hierarchical}. 

As we point out, both RL and MCTS have weaknesses when applied individually to complex decision-making problems in non-stationary environments. 
We argue that a hybrid decision-making approach that integrates RL and online planning can combine their strengths while mitigating their individual weaknesses in non-stationary environments. The intuition is that if the environment has not changed too much between when an optimal policy was learned and when a decision needs to be made, the policy can still provide useful information for decision-making. Similar hybrid approaches have been explored and have achieved state-of-the-art performance in many domains. 
For example, AlphaZero~\citep{davidsilver2018} integrates a policy and value network within a modified MCTS, and has achieved state-of-the-art performance in games such as Go. However, to the best of our knowledge, such a hybrid approach has not been developed specifically for non-stationary environments.

In this paper, we show how hybrid approaches can combine offline learning and online search for decision-making in non-stationary environments. We present a novel hybrid decision-making approach, called \textit{Policy-Augmented Monte Carlo tree search} (PA-MCTS). Our approach is remarkably simple---it combines a policy's action-value estimates with the returns generated by MCTS \textit{without changing either of the two approaches}, i.e., the combination occurs entirely outside the online search tree. Specifically, we make the following contributions:
\begin{enumerate}
    \item Conceptually, our core contribution is the idea that offline planning and online search can be combined for decision-making in non-stationary environments. 
    \item We present two algorithms to operationalize the proposed idea. The first approach, our core algorithmic contribution, is a novel algorithm that combines a (relatively) stale policy with MCTS after the search, i.e., completely outside the tree. We present several theoretical results for the proposed approach. Second, we show how existing hybrid approaches, e.g., AlphaZero can also be used for decision-making in non-stationary environments.
    \item We validate our approach using four open-source environments from OpenAI Gym and compare it with other state-of-the-art approaches. We show that PA-MCTS results in two distinct advantages---first, given a specific computational budget, our framework converges to significantly better decisions than standard MCTS; and second, online search makes our approach significantly more robust to environmental changes than standard state-of-the-art approaches. 
\end{enumerate}

%% file: background.tex
\section{Background} \label{sec:related_work}


\textbf{Motivating Environments:} 
Consider the problem of proactively allocating ambulances in a city in anticipation of accidents, modeled as an MDP~\citep{mukhopadhyay2018decision}. Now, consider that the city experiences unexpected congestion, changing the underlying MDP's transition function. First responders cannot wait to re-train a new policy, and a sub-optimal policy can waste invaluable time to respond to emergency situations. Abstraction of such a problem can be modeled in simpler settings such as toy environments in OpenAI Gym~\citep{brockman2016openai}. For example, in the cartpole environment, a rigid pole is attached to a cart via a hinge. The cart can move freely on a bounded, frictionless horizontal track and the agent (i.e., the controller) can apply a force to the cart parallel to the track in either direction. The goal of the controller is to balance the pole as long as possible. Non-stationarity can be induced by varying the gravitational constant or the mass of the cart. Similarly, the Frozen Lake environment involves an agent trying to cross a frozen lake from a start position to a goal position (pre-determined) without falling into any holes (we show a schematic of the environment in \cref{figure:frozenLake}). However, the slippery nature of the surface results in the agent not always moving in the intended direction.
Here, non-stationarity can be induced by changing the coefficient of friction on the surface, thereby changing how the agent moves. In all settings, we explore how the agent can quickly adapt to the updated environment.

\textbf{Monte Carlo Tree Search:} Our approach is based on Monte Carlo Tree Search (MCTS), an anytime search algorithm that builds a search tree in an incremental and asymmetric manner~\citep{coulom2006efficient, kocsis2006bandit, browne2012survey}. 
MCTS represents environment states by nodes, with the current state being the root node. Actions that result in a transition from one state to another are represented as edges between nodes.
The fundamental idea of MCTS is to bias the search toward actions that appear promising. To determine which actions are promising, MCTS uses a model $\mathcal{M}_{t}$ of the environment at the current decision epoch $t$ to simulate a \textit{rollout} (commonly a randomly sampled possible future trajectory) to the end of the planning horizon.
A \textit{tree policy}, such as the commonly used Upper Confidence bound for Trees (UCT) algorithm~\citep{kocsis2006bandit}, then uses the averaged returns of these rollouts as estimates $\overline{G}_{t}(s, a)$ for the value of each action $a$ at each state $s$, and biases the exploration of future rollouts towards higher estimated returns. MCTS is proven to converge to the optimal action given infinite time~\citep{kocsis2006bandit}. However, the number of iterations required can become impractical as the state-action space of the environment grows.

\textbf{AlphaZero:} One of the most well-known approaches that combine the strengths of MCTS-based planning and deep learning is  AlphaZero~\citep{silver2018general}, which has shown state-of-the-art performance in games such as Chess and Go. AlphaZero consists of a neural network to estimate the value of states and MCTS is used as a means for policy improvement. While AlphaZero was not designed as a framework for decision-making in non-stationary environments, we hypothesize that it is particularly suited to such environments. We refer readers to the seminal work by \citet{silver2018general} for a comprehensive description of the AlphaZero framework. While we show experimental results against AlphaZero as a baseline, we reiterate that showing that AlphaZero can be used as a decision-making tool for non-stationary environments is one of our contributions. 

%% file: Problem.tex
\section{Markov Decision Processes in Non-Stationary Settings} \label{sec:problem}

Markov decision processes (MDP) provide a general framework for sequential decision-making under uncertainty. An MDP can be defined by the tuple $(\mathcal{S}, \mathcal{A},P(s,a),r(s,a))$, where $\mathcal{S}$ is a finite state space, $\mathcal{A}$ is a discrete action-space, $P(s' \mid s,a)$ is the probability of reaching state $s'$ when taking action $a$ in state $s$, and $r(s,a)$ is the scalar reward when action $a$ is taken in state $s$.
The goal of an agent is to learn a policy $\pi$ that maps states to actions (or, more generally, states to a distribution over actions) that can maximize a specified utility function. Typically, the utility function is simply the expected cumulative reward, also called the \textit{return}, which is defined as $\mathbb{E}_{P}[\sum_{t=0}^{\infty}\gamma^t r(s_t, \pi(s))]$, where $\gamma$ denotes the discount factor that weighs immediate rewards more than future rewards. We are interested in settings where the environment with which the agent interacts changes over time. Specifically, we consider an agent that is \textit{trained} on a specific task given a particular setting, 
but is \textit{executed} in an environment where the underlying MDP's transition function has been modified; e.g., consider the problem of emergency response described in \cref{sec:related_work}). 


We point out that our interest lies in problem settings where \textit{learning} a new policy immediately is infeasible in practice. In complex real-world tasks, learning a new policy takes time, and decisions must be made as the policy is being updated based on the new environmental conditions.
We are interested in understanding how agents can optimize decision-making during such delays, i.e., between when a change is detected in the environment and when the agent learns a new (near-optimal) policy. One way to model such a problem is a non-stationary Markov decision process (NSMDP)~\citep{lecarpentier2019non}. 
An NSMDP, as defined by Lecarpentier and Rachelson~\citeyearpar{lecarpentier2019non}, can be viewed as a stationary MDP where the state space is extended with a temporal dimension. Depending on whether the task is episodic and whether the agent is allowed to explore along the temporal axis, this enhancement can be trivial or difficult~\citep{lecarpentier2019non}.  

The rate of change in the transition function is typically bounded in an NSMDP by assuming Lipschitz continuity~\citep{lecarpentier2019non}, which assumes that the function changes smoothly over time. However, some situations can cause abrupt, large changes to the dynamics of the environment. We propose a new approach to bound the non-stationarity of an MDP by instead considering an upper bound on the total change in the transition function between training and execution; we refer to such a setting as \emph{transition-bounded non-stationary Markov decision processes} (T-NSMDP). 
Consider the transition probability function $P_{t}(s'\mid s, a)$, where the subscript $t$ denotes the time step under consideration. Now, consider that the environment undergoes \textit{some} change between time steps $0$ and $t$. We assume that: 

\begin{equation}
    \forall s, a: \sum_{s'\in S} \left| P_{t}(s'\mid a, s) - P_{0}(s'\mid a, s) \right| \leq \eta
    \label{eq:assumption:transition_changes}
\end{equation}
where $t\in\mathcal{T}$ (i.e., some point in time after the original policy was learned), and $\eta \in \mathbb{R}^+$ is a scalar bound. In other words, the total change in the transition function (cumulated over all states $s\in \mathcal{S}$ and actions $a\in \mathcal{A}$) \hb{It's cumulated over all possible next states, isn't it? Confusing sentence} between the updated environmental conditions and the original environmental conditions is bounded by $\eta$. 
We also point out that while our problem definition is agnostic to whether the change is continuous or discrete (we could define an analogous setting for continuous-time MDPs), our algorithm only tackles discrete changes for now.


\begin{figure}[ht]
\centering
\begin{tikzpicture}
\centering
  \draw (0,0) grid (4,4); 
  \filldraw[fill=white!30] (0,0) rectangle (4,4);
  \foreach \x/\y/\label in {2/1/A,3/0/G, 1/2/H, 3/2/H, 0/0/H}
    \node at (\x+0.5, \y+0.5) {\Huge\textbf{\label}};
  \foreach \x in {0,1,2,3,4}
    \node[anchor=north] at (\x,0) {\x};
  \foreach \y in {0,1,2,3,4}
    \node[anchor=east] at (0,\y) {\y};
  \node[align=center,font=\bfseries] at (2,4.5) {Frozen Lake Scenario};
  \draw[step=1, black, thick] (0,0) grid (4,4);
\end{tikzpicture}
\caption{The Frozen Lake environment involves an agent trying to cross a frozen lake from a start position to a goal position (pre-determined) without falling into any holes}
\label{figure:frozenLake}
\end{figure}
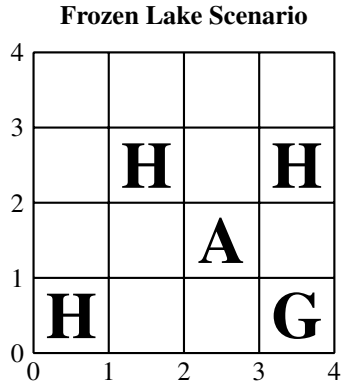

We demonstrate how \cref{eq:assumption:transition_changes} can be applied in practice using the Frozen Lake environment described earlier. At each state, the agent has four actions, i.e., the direction in which it chooses to move. Consider that in the original environment at time $0$, the lake is only marginally slippery such that the agent can always go in the direction it intends. Let us consider an arbitrary state $s_{(2, 1)}$ which denotes that the agent is in cell $(2,1)$ (refer to the \cref{figure:frozenLake} on the left where each cell is marked with a number). In the original environment, if the agent seeks to go right (takes action $a_{\text{right}}$), then $P_0(s_{(3,1)} \mid a_{\text{right}}, s_{(2,1)}) = 1$, i.e., the agent always moves right if it intends to do so. \hb{We're describing a different default here than what we described to be the default above. Maybe leave out the default above and speak simply of different ways the environment could work?}
Now, at time $t$, consider that lake becomes more slippery, and the agent may not always move in the intended direction due to the slippery nature of the surface. Let us assume that the agent moves in the intended direction with a probability of $1/3$ and in each of its perpendicular directions with the same probability (i.e., from cell (2,1), if the agent seeks to go right, it can now end up in cells (3,1), (2,0), or (2,2) with equal probability). In that case, by equation \cref{eq:assumption:transition_changes}, we can quantify the \textit{amount of change}\footnote{we show this for one particular state as an example, the same bound applies across all state-action pairs.} to be $\mid P_t(s_{(3,1)} \mid a_{\text{right}}, s_{(2,1)}) - P_0(s_{(3,1)} \mid a_{\text{right}}, s_{(2,1)}) \mid$ + $\mid P_t(s_{(2,0)} \mid a_{\text{right}}, s_{(2,1)}) - P_0(s_{(2,0)} \mid a_{\text{right}}, s_{(2,1)}) \mid$ + $\mid P_t(s_{(3,3)} \mid a_{\text{right}}, s_{(2,1)}) - P_0(s_{(3,3)} \mid a_{\text{right}}, s_{(2,1)}) \mid$ + $\mid P_t(s_6 \mid a_{\text{right}}, s_{(2,1)}) - P_0(s_{(2,2)} \mid a_{\text{right}}, s_{(2,1)}) \mid$ = $1.33$. A similar bound can be computed for the Cartpole environment where the agent is trained in a setting with a certain mass, but the mass of the cart changes in the updated environment.



\textbf{Hypothesis:} Having described an example of how the bounds in a T-NSMDP can be computed, we now present our key hypothesis. Let $s_0 \in \mathcal{S}$ and $a_0 \in \mathcal{A}$ denote the initial state and an action (respectively) at time step 0. Consider the action-value function $Q$, represented as $Q^{\pi}(s,a) = \mathbb{E}_{P}\{ R(s_0, a_0) + \sum_{t=1}^{\infty} \gamma^t R(s_t, a_t = \pi(s_t))\}$, which measures the value of a state $s$ and an action~$a$ under policy $\pi$. Our key hypothesis is that with small changes in the transition function, the $Q$ function under an optimal policy does not change much, i.e., ``good'' actions remain valuable, and ``bad'' actions do not suddenly become promising. It is trivial to see that there are exceptions to our hypothesis and its rather simplified explanation; if the $Q$ function stayed the same as before, there would not be any need for the agent to adapt to changes in the environment. However, we show that the change in $Q$ is bounded with respect to the change in $P$: 

\begin{theorem}
    \label{prop:q_bound}
    If \ $\forall s, a \colon \sum_{s'\in S} \left| P_{t}(s'\mid a, s) - P_{0}(s'\mid a, s) \right| \leq \eta$, and \\ $\forall s, a   \colon |r(s, a)| \leq R$, and the discount factor $\gamma < 1$,  
    then $|Q_{0}^{\pi^{*}_{0}}(s,a)-Q_{t}^{\pi^{*}_{t}}(s,a)| \leq \epsilon \  \forall s, a$, where \ $\epsilon = \frac{\gamma \cdot \eta \cdot R}{(1 - \gamma)^2}$ (The proof is presented in the appendix).
    \vspace{-0.2mm}
\end{theorem}
\cref{prop:q_bound} considers two environments, one at time $0$ and the other at time $t$; then, under the assumptions that \cref{eq:assumption:transition_changes} holds, the rewards are bounded, and $\gamma<1$, the difference in the optimal Q values (in the old and new environment) is also bounded. While this bound is admittedly loose and can be large in many settings, it serves as the basis for algorithmic decision-making in non-stationary settings if the environment has not changed much. Specifically, if the optimal Q has not changed significantly, then we can leverage it to inform decision-making even when the environment has changed. \hb{To be fair, for a typical $\gamma$ close to 1, $\epsilon$ gets huge...?}



\textbf{Problem Statement:} Our goal is to find optimal actions (i.e., actions that maximize the temporally discounted sum of future rewards) at execution time $t$ assuming the following: (1) we are given an optimal action-value function $Q_{0}^{\pi^{*}_{0}}(s,a)$ for time step $0$, which might have been learned under different environmental conditions than those at execution time $t$; (2) any such change in the transition probabilities between when $Q_{0}^{\pi^{*}_{0}}(s,a)$ was learned and time $t$ are bounded by $\eta$ (\cref{eq:assumption:transition_changes}); (3) we are given a black-box simulator that is updated to accurately model the environment at execution time $t$ (we relax this assumption in our experiments); and (4) there is a limited computational budget at execution time that prevents learning an optimal policy.




%% file: approach.tex
\section{Policy Augmented Monte Caro Tree Search}\label{sec:approach}

Recall that MCTS is proven to converge to the optimal action given infinite time~\citep{kocsis2006bandit}. The convergence also holds true for an environment that has changed---this consideration is actually meaningless for a purely online approach such as MCTS, assuming that the generative model that is used to build the tree has been updated based on environmental changes. On the other hand, a trained policy has exactly the opposite advantage (and disadvantage): while it can be invoked in constant time during decision-making, it cannot accommodate environmental changes without re-training, which is computationally expensive. We raise the following question: \textit{can the advantages of online search procedures (e.g., MCTS) be combined with those of policy learning (e.g., reinforcement learning) to tackle discrete changes in the environment?}

Our approach presents a natural solution to this question. \textit{Policy-Augmented Monte Carlo Tree Search} (PA-MCTS) addresses this challenge by integrating an online search with $Q$-values learned on the environment at an earlier decision epoch, even if the environment has changed. Rather than selecting an action based on the highest expected return estimated by the online search, PA-MCTS instead chooses the action that maximizes a convex combination of the previously learned $Q$-values and the MCTS estimates~$\overline{G}$:
\begin{equation}
    \argmax_{a \in \mathcal{A}_s} \ \ \ \alpha Q_{0}^{\pi^{*}_{0}}(s, a) + (1 - \alpha)\overline{G}_{t}(s, a) \label{eq:pamcts_selection}
\end{equation}
where $Q_{0}^{\pi^{*}_{0}}(s, a)$ is the optimal\footnote{In principle, we do not require the optimal $Q$-function. As we show in the experiments, an approximation also works well in practice.} $Q$-function previously learned by the decision agent. Note that this combination happens entirely \textit{outside the tree}.\footnote{We also explored approaches to use the learned Q-values inside the tree, but the proposed approach performed at least as well and simplifies (to some degree) the theoretical analysis.} The hyper-parameter $\alpha$, chosen such that $0 \le \alpha \le 1$, controls the tradeoff between the learned $Q$-values and the expected returns generated through MCTS estimates: if $\alpha = 1$, PA-MCTS reduces to the standard Q-learning action selection policy using the previous $Q$ values: $\argmax_{a \in \mathcal{A}_s} Q_{0}^{\pi^{*}_{0}}(s, a)$. If $\alpha = 0$, it reduces to standard MCTS. We essentially seek to balance the dichotomy between using low-variance but biased estimates through $Q_{0}^{\pi^{*}_{0}}$ (generated using an older environment) and potentially high-variance but unbiased estimates through $\overline{G}_{t}$ (generated using the current environment). \hb{This could be confusing for the reader, because the Q-values are called accurate but aren't accurate anymore, while the MCTS approximation is valid for the current environment. Would it be better to say that the Q-values have low variance but are biased through the non-stationarity, and the MCTS approximation has high variance but is unbiased?}
If $\alpha \in (0, 1)$, then both estimates are considered. Note that we refer to the action-value from the $Q$-function as ``estimates'' due to the change in the environment. We hypothesize that when the error in $Q$-values is bounded by $\epsilon$ (as described in \cref{prop:q_bound}), $Q_{0}^{\pi^{*}_{0}}$ likely embeds useful information about the updated environment. The PA-MCTS algorithm essentially uses \cref{eq:pamcts_selection} for decision-making. We present the algorithm in the appendix.

%% file: proof.tex
\subsection{Theoretical Analysis}\label{sec:proofs}

We prove three properties of PA-MCTS: (1) the conditions under which PA-MCTS will return the optimal one-step action, (2) the conditions under which PA-MCTS will choose an action with a higher estimated return than either MCTS or selection using $Q$-values $Q_{0}^{\pi^{*}_{0}}$, and (3) a bound on the total deviation of the expected return from an optimal (updated) policy when following PA-MCTS. We begin by defining some additional notation and detailing our assumptions. 

Let $a_{t}^{*} := \argmax_{a \in \mathcal{A}_s} Q^{\pi^{*}_{t}}(s, a)$ be the optimal action at an arbitrary time step~$t$. Let $a'_{t} := \argmax_{a \in \mathcal{A}_s \setminus \{a^{*}_{t}\}}  Q^{\pi^{*}_{t}}_{t}(s, a)$ denote the second best action at an arbitrary time step~$t$ (we assume that there are no ties in $Q$-values for any actions at a given state). Let $\psi_{t}(s) := Q^{\pi^{*}_{t}}_{t}(s,a_{t}^*) - Q^{\pi^{*}_{t}}_{t}(s,a_{t}')$; for a state $s$, $\psi_{t}(s)$ denotes the difference in $Q$ values when taking actions $a_{t}^*$ and $a_{t}'$ at time $t$ and following the optimal policy $\pi^{*}_{t}$ thereafter.\footnote{Recall that the optimal policy is indexed with the time step as the environment might change in our setting, i.e., $\pi^{*}_{t}$ denotes the optimal policy given the environmental conditions at time step $t$.} Finally, while MCTS is guaranteed to converge to the optimal expected returns for a given state and action given infinite time, actions must be taken after a limited time in practice. Let $\delta$ denote the bound on the error of the values estimated by MCTS when it is stopped, i.e., $|Q_{t}^{\pi^{*}_{t}}(s,a) - \overline{G}_{t}(s, a)|_{\infty} \leq \delta \ \ \ \ \forall  \ \ t \in \mathcal{T}$.

We first present Theorem~\ref{prop:optimal_onestep_action}, which describes the conditions with respect to $\epsilon$, $\delta$, and $\psi_{t}(s)$ under which PA-MCTS returns the optimal one-step action. 

\begin{theorem}
\label{prop:optimal_onestep_action}
If $\alpha\epsilon + (1-\alpha)\delta \leq \frac{\psi_{t}(s)}{2}$, PA-MCTS is guaranteed to select the optimal action at time step $t$ in state $s$. (The proof is presented in the  appendix).
\end{theorem}


Theorem~\ref{prop:optimal_onestep_action} essentially shows that PA-MCTS is guaranteed to choose the optimal action if the sum of $\alpha$-weighted errors is low enough that the decision agent can differentiate between the best and the next best action. While $\psi_{t}(s)$ is not observable at decision time, we can find the relationship between $\psi_{t}(s)$ and $\psi_{0}(s)$:
\begin{corollary}
\label{lem:psi_relation}
$\psi_{t}(s) \leq \psi_{0}(s) + 2\epsilon$
\end{corollary}
Using \cref{lem:psi_relation}, we substitute $\psi_{t}(s)$ in terms of $\psi_{0}(s)$ (which can be computed at decision time using known $Q^{\pi^{*}_{0}}_{0}$ values) in Theorem~\ref{prop:optimal_onestep_action}. Then, we can determine when PA-MCTS will select the optimal action using information that can be computed at decision time:
\begin{corollary}
\label{lem:psi_0_bound}
If $\alpha\epsilon + (1-\alpha)\delta \leq \frac{\psi_{0}(s)}{2} + \epsilon$, PA-MCTS is guaranteed to select the optimal action at decision epoch $t$. 
\end{corollary}
Using $0 \le \alpha \le 1$, \cref{lem:psi_0_bound} can be rearranged to solve for $\alpha$:
\begin{corollary}
\label{lem:alpha_bound}
PA-MCTS will choose the optimal one-step action if
\begin{equation}
    \begin{cases}
        \frac{-\delta}{\epsilon - \delta} \leq \alpha \leq \frac{\psi_{0}(s)}{2(\epsilon - \delta)} + 1 &\text{if $\epsilon > \delta$}\\
        \frac{-\delta}{\epsilon - \delta} \ge \alpha \ge \frac{\psi_{0}(s)}{2(\epsilon - \delta)} + 1 &\text{if $\epsilon < \delta$.}
    \end{cases} \nonumber
\end{equation} 
\end{corollary}
Therefore, given $\epsilon$ and $\delta$, a decision agent can determine what values of the hyperparameter $\alpha$ (if any) would guarantee that PA-MCTS chooses the optimal action at decision time. 

\cref{prop:optimal_onestep_action} guarantees the conditions under which PA-MCTS will choose the optimal action. We now seek to answer a more practical question: when does PA-MCTS choose a one-step action with a higher $Q^{\pi^{*}_{t}}_{t}$ value than MCTS or the learned policy in isolation? First, we list conditions under which PA-MCTS chooses a better action than pure MCTS:
\begin{prop}
\label{prop:pamcts_better_mcts}
If PA-MCTS and MCTS choose different actions, PA-MCTS's chosen action will have a higher $Q^{\pi^{*}_{t}}_{t}$ value than MCTS if $2\epsilon\ \leq \zeta^{m}_{t}$, where\footnote{the superscript $m$ in $\zeta^{m}_{t}$ represents the bounds with respect to MCTS and is not an index.} $\zeta^{m}_{t} := (Q^{\pi^{*}_{0}}_{0}(s, \tilde{a}) + \overline{G}_{t}(s,  \tilde{a})) - (Q^{\pi^{*}_{0}}_{0}(s, a_m) + \overline{G}_{t}(s,  a_m))$, $\tilde{a} := \argmax_{a \in \mathcal{A}_s} \ \ \ \alpha Q^{\pi^{*}_{0}}_{0}(s, a) + (1-\alpha)\overline{G}_{t}(s,  a)$, and $a_m := \argmax_{a \in \mathcal{A}_s} \overline{G}_{t}(s,a)$. (The proof is presented in the  appendix).
\end{prop}

Since $\zeta^{m}_{t}$ is constructed from terms that can be computed at decision time, Proposition \ref{prop:pamcts_better_mcts} can be used by the decision agent to determine if it should use PA-MCTS or MCTS when they choose different actions at decision time. 
A similar statement can be made regarding action selection using $Q^{\pi^{*}_{0}}_{0}$.
\begin{prop}
\label{prop:pamcts_better_policy}
If the action chosen by PA-MCTS is different than $\argmax_{a}Q^{\pi^{*}_{0}}_{0}(s,a)$ at state $s$, then PA-MCTS's chosen action will have a higher $Q^{\pi^{*}_{t}}_{t}$ value than selection through $Q^{\pi^{*}_{0}}_{0}$-values if $2\delta\ \leq \zeta^{q}_{t}$ where\footnote{Again, the superscript $q$ only denotes bounds with respect to the $Q$-values and is not an index.} $\zeta^{q}_{t} := (Q^{\pi^{*}_{0}}_{0}(s, \tilde{a}) + \overline{G}_{t}(s, \tilde{a})) - (Q^{\pi^{*}_{0}}_{0}(s, a_g) + \overline{G}_{t}(s,  a_g))$, $\tilde{a} := \argmax_{a \in \mathcal{A}_s} \alpha Q^{\pi^{*}_{0}}_{0}(s, a) + (1-\alpha)\overline{G}_{t}(s, a)$, and $a_g := \argmax_{a \in \mathcal{A}_s} Q^{\pi^{*}_{0}}_{0}(s,a)$. (The proof is presented in the  appendix).
\end{prop}
Using Propositions~\ref{prop:pamcts_better_mcts} and~\ref{prop:pamcts_better_policy}, we can determine the conditions under which PA-MCTS chooses an action that is better than either of its constituent policies. The above propositions consider only one decision epoch. However, typically, training a new policy takes time, which might consist of several decision epochs. Therefore, we compute the total error in the expected return when following PA-MCTS compared to an optimal (updated) policy.
\begin{theorem}
\label{theorem:return_error}
When PA-MCTS is used for sequential decision making, the maximum difference between the return from an optimal policy and the return from following PA-MCTS is at most $\frac{2(\alpha\epsilon - \alpha\delta + \delta)}{1-\gamma}$. (The proof is presented in the  appendix).
\end{theorem}

%% file: experiments_ayan.tex
\subsection{Experiment Setting}\label{sec:exp}

\textbf{Environments:} We use the following four open-source environments from OpenAI Gym~\citep{brockman2016openai} to validate our approach:
\begin{enumerate}
    \item \textbf{Cart Pole}: In this classic control problem, a pole is attached by an un-actuated joint to a cart, which moves on a frictionless track. A pendulum is placed upright on the cart. The agent's goal is to balance the pole by applying forces either to the left or the right direction on the cart. For the cart pole environment, we set our original environment with gravitational constant g as $9.8\,\text{m/s}^2$ and the mass of the pole as $0.1 \,\text{kg}$. Both parameters are then varied to induce non-stationarity in the state-action transition function. 
    \item \textbf{Frozen Lake}: In the frozen lake problem, an agent must traverse from a starting position to the goal position by avoiding holes. The agent moves on a slippery surface (we describe the setting in section~\ref{sec:problem}). We introduce non-stationary by making the surface more slippery; in our experiments, we refer to a problem instance of the frozen lake environment by the vector $p_1, p_2, p_3$, where $p_1$ denotes the probability with which the agent moves in the intended direction, and $p_2$ and $p_3$ denote the probabilities of moving in directions perpendicular to the intended direction. For the frozen lake environment, we treat the ``original environment'' as one with perfect agent movement, i.e., deterministic movement.
    \item \textbf{Cliff Walking}: In the cliff walking environment, an agent must find a path from the start position to the goal position from start to goal while avoiding falling off a cliff. While the setting is similar to the frozen lake environment, here, the agent incurs a penalty for each step that it takes, thereby presenting the explicit dichotomy of choosing between an optimal path (with respect to discounted rewards) and a safe path. Similar to the frozen lake environment, the original environment involves deterministic agent movement. 
    \item \textbf{Lunar Lander}: This environment presents a rocket trajectory optimization problem. The agent must control a rocket and land safely on a known landing pad. While landing outside the pad is possible, the agent incurs a penalty. The agent's action at each time step involves firing one of the three engines (main, right, and left) at full throttle or not doing anything. Here, the original environment does not involve any wind. We induce wind to simulate non-stationarity; the agent, therefore, must adjust its actions according to the power of the wind.
\end{enumerate}

\input{results}

\noindent \textbf{Baseline approaches and implementation:} We implement AlphaZero, a reinforcement learning (RL) agent using a Double Deep Q-Network (DDQN)~\citep{van2016deep}, and standard MCTS as our baseline approaches. We reiterate that we do not treat AlphaZero as a \textit{baseline} per se since our broader conceptual idea of using hybrid approaches to combat non-stationarity subsumes it. Nonetheless, we compare PA-MCTS with AlphaZero, and show that our proposed approach outperforms AlphaZero. We implement MCTS and PA-MCTS using UCT~\citep{kocsis2006bandit} as the tree policy. Our implementation is in Python~\citep{van1995python} and available at \texttt{https://github.com/scope-lab-vu/PAMCTS}\\
\textbf{Hardware:} We run experiments on the Chameleon testbed~\citep{keahey2020lessons} on 4 Linux systems with 32-96 logical processors and 528 GB RAM. We train AlphaZero using the Google Cloud Platform on a Linux system with 128 logical processors and 528 GB RAM.\\
\textbf{Hyper-Parameters:} We implement the neural networks in TensorFlow~\citep{tensorflow2015-whitepaper}. For the RL agent, the Double DQN has 3 hidden layers with ReLU activation, and the output layer has a linear activation. We use a Boltzmann Q Policy and train the network with the Adam Optimizer\citep{kingma2014adam} with a learning rate of $10^{-3}$ for 300,000 steps while minimizing the mean absolute error of the network. For AlphaZero, we use 5 hidden layers with ReLU activation and appropriate output layers (e.g., softmax for the policy and dense layer for the value function). For both PA-MCTS and AlphaZero, we set the discount factor as 0.99, maximum tree depth as 500 (though this is rarely reached), and the exploration parameter $c$ = 50 (after tuning on holdout data). A more detailed description of the hyper-parameters is presented in the appendix. \hb{Link to that? The number of neurons would be interesting}

\subsection{Results}

Comprehensive results for all environments and all settings we use to induce non-stationarity are presented in Table~\ref{tab:results}. Below, we dive deeper into the cart pole environment to analyze the results in depth and evaluate the nuances of PA-MCTS. Similar analysis for the other environments can be found in the appendix.


\begin{figure}[!htb]
	\centering
	\begin{subfigure}{.5\textwidth}
		\centering
		\includegraphics[width=\linewidth]{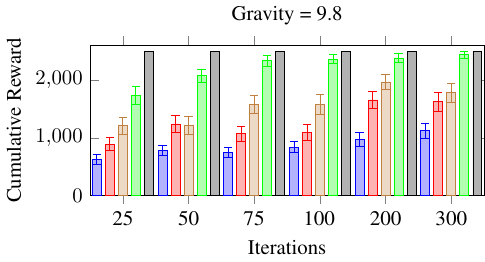}
	\end{subfigure}%
	\begin{subfigure}{.47\textwidth}
		\centering
		\includegraphics[width=\linewidth]{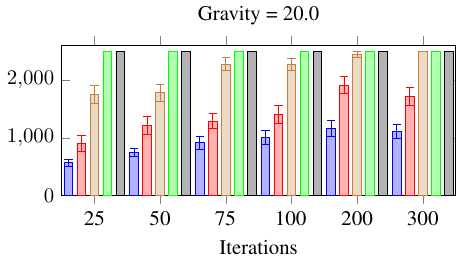}
	\end{subfigure}
	\begin{subfigure}{.5\textwidth}
		\centering
		\includegraphics[width=\linewidth]{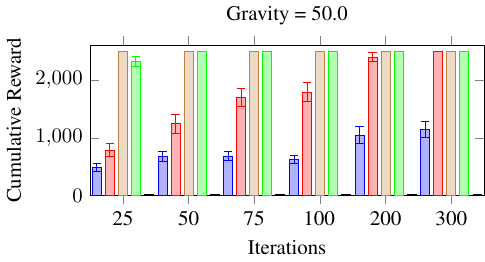}
	\end{subfigure}%
	\begin{subfigure}{.47\textwidth}
		\centering
		\includegraphics[width=\linewidth]{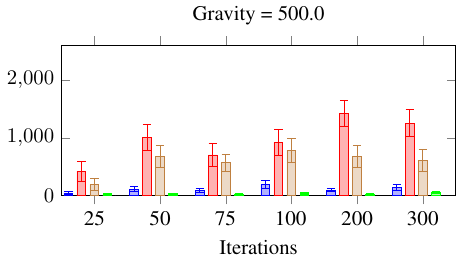}
	\end{subfigure}
	\begin{subfigure}{\textwidth}
		\centering
		\includegraphics[width=0.8\linewidth]{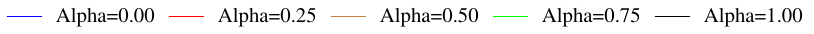}
		\caption*{} 
	\end{subfigure}
	
	\caption{We show the cumulative reward obtained by PA-MCTS, standard MCTS (\(\alpha = 0\)), and DQN (\(\alpha = 1\)) for different environmental changes. Note that as the environment changes, DQN achieves close to 0 rewards. PA-MCTS outperforms both baseline approaches.}
	\label{fig:pamcts_cartpole_gravity}
\end{figure}

\textbf{Stationary setting:} We train both the AlphaZero network and the DDQN with $g=9.8\,\text{m/s}^2$ and $m=1.0\,$kg and evaluate all the approaches in the same environmental settings. \ava{The meaning of `evaluate all the approaches in the same environmental settings' is not clear to me.}
We present PA-MCTS and DDQN results in \cref{fig:pamcts_cartpole_gravity} (different color lines represent different $\alpha$ used by PA-MCTS) and the comparison with AlphaZero in \cref{figure:az_gravity} (the left-most sub-graph shows the stationary setting). 
\ava{Figure 2 only has dqn. Rewrote to  fix; check it for correctness.} \hb{The figure has only one row. Also, the legend is currently unreadable} We observe that under stationary conditions, both AlphaZero and the DDQN (which is a special case of PA-MCTS with $\alpha=1.0$) outperform standard MCTS and PA-MCTS with $\alpha\in\{0.25, 0.5, 0.75\}$. This result is expected, as offline learning is able to leverage potentially thousands of training episodes to learn an optimal policy. However, we do note that PA-MCTS (for $0<\alpha<1$) is able to leverage the information stored in the policy to converge significantly faster than standard MCTS. We compute each result by averaging across 30 samples. \\
\textbf{Non-stationary setting:} To add non-stationarity to the cart pole environment, we modify $g$ from $9.8\,\text{m/s}^2$ to $20, 50,$ and $500\,\text{m/s}^2$ (we present results in \cref{fig:pamcts_cartpole_gravity}; additional experiments which modify the mass of the cart are shown in the appendix). We have several important findings. First, we observe as the change in the environment increases, the performance of the DDQN policy in isolation (trained on the original environment) degrades as hypothesized---in fact, it achieves almost 0 utility for $g \in \{50, 500\}$. Second, we again observe that PA-MCTS converges significantly faster than standard MCTS for all values of $g$ (with appropriate $\alpha$); it actually achieves optimal performance except when $g=500\,\text{m/s}^2$. 
\ava{I might be misinterpreting the plots, but this analysis looks incorrect. My observations: 1) The performance of DDQN AND AlphaZero degrade as the environment changes. 2) PA-MCTS converges significantly faster than standard MCTS for all values of $g$ (with appropriate $\alpha$).} 
The reader might wonder about how to choose the optimal $\alpha$ and the performance of AlphaZero, which we show below. \ava{why is the alphazero comparison in a different subsection? If we want to do this, we should indicate this section is focused on DDQN/MCTS in the title.}\\
\begin{figure}
\centering
\includegraphics[width=0.5\linewidth]{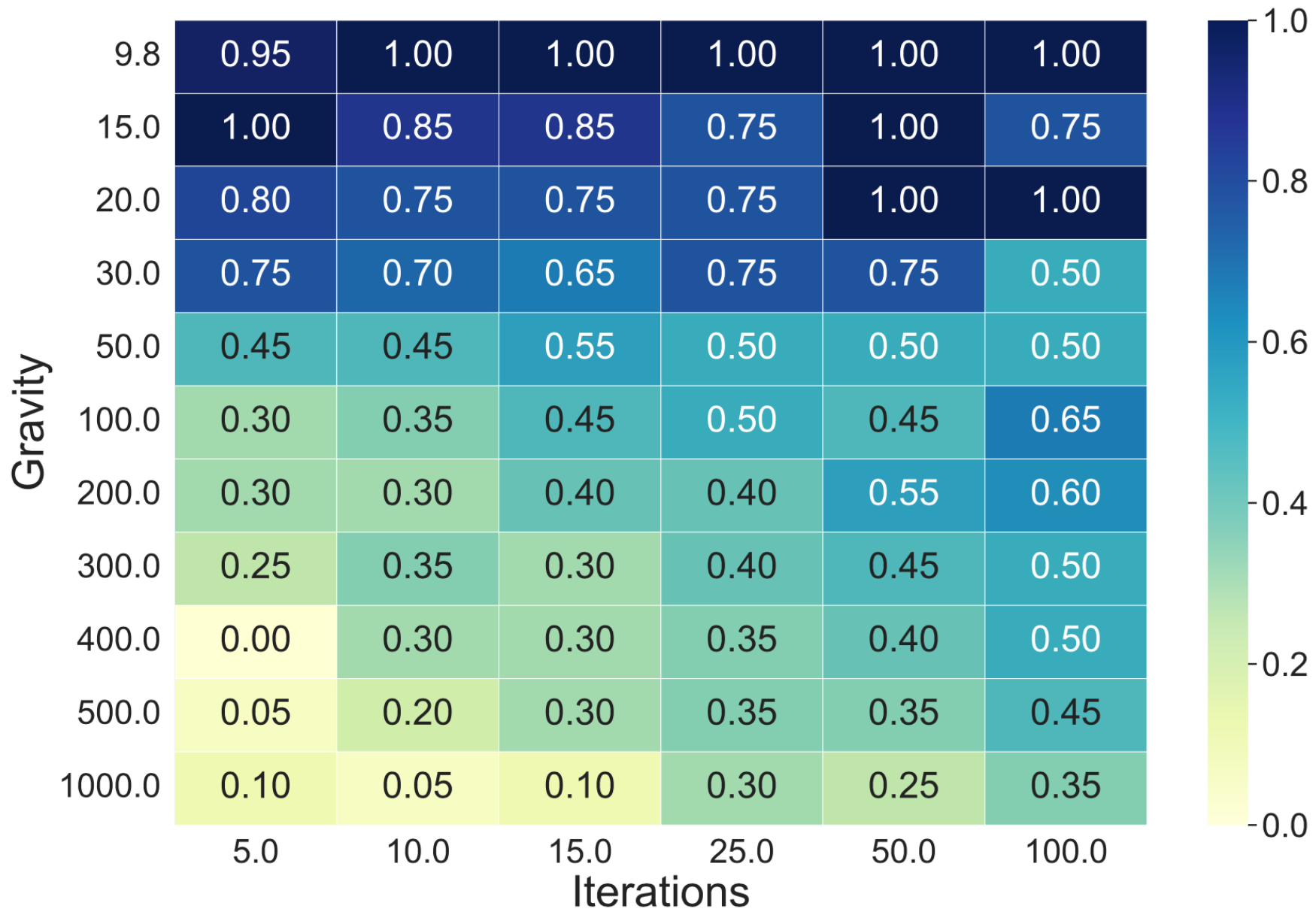}
\caption{We show the optimal $\alpha$ for varying MCTS iterations and environmental conditions.}
\label{figure:heatmap_g}
\end{figure}
\begin{figure}
\centering
\includegraphics[width=0.6\linewidth]{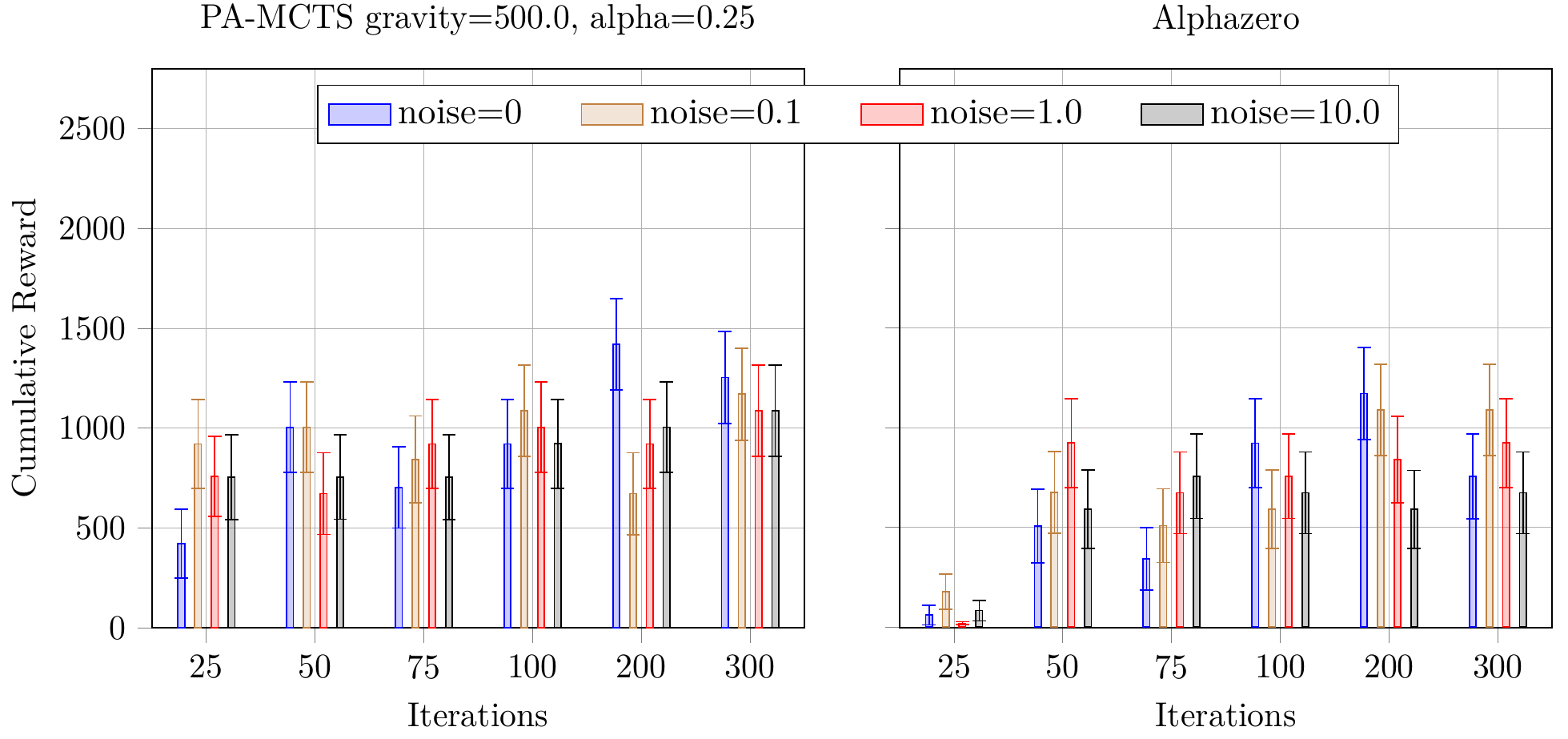}
\caption{The performance of PA-MCTS and AlphaZero under varying levels of noise with $g=500\,\text{m/s}^2$. PA-MCTS outperforms AlphaZero in most settings.}
\label{figure:bar_plots}
\end{figure}
\textbf{Choosing the optimal $\alpha$:} During execution, the agent must choose actions relatively quickly. Therefore, even when updated generative models of the system are available, it is unlikely that the agent would be able to afford the time to compute extensive simulations across a range of $\alpha$ values to select the optimal one. We provide an empirical approach to alleviate this concern. First, we run simulations for various iteration counts of MCTS in a wide array of environmental conditions. For each setting, we show the optimal $\alpha$ in \cref{figure:heatmap_g}. For example, the top rows of the heatmap show that when the environment has not changed much, it is better for the agent to rely more on the trained policy (i.e., $\alpha$ is typically above 0.5). On the other hand, consider the bottom-right cell in the heatmap, which shows that if the environment changes a lot, the agent must rely more on the online search, as expected. We point out another interesting finding---it turns out that PA-MCTS, with limited MCTS iterations, is able to approximate the optimal choice of $\alpha$ fairly well, i.e., observe that the column with iterations=25 (or even with iterations=10) is almost similar to the column with iterations=100. We use this finding to guide the selection of $\alpha$. We hypothesize that the agent can use the updated generative models to perform a hyper-parameter sweep of $\alpha$ with a small number of MCTS iterations; the sweep can be run in parallel for each $\alpha$ and in our setting takes only 80 seconds (much smaller than re-training the policy, as we show below). Then, for execution, the agent can choose the $\alpha$ value that provides the highest expected discounted return.\\

\begin{figure}[!htb]
	\centering
	\begin{subfigure}{.5\textwidth}
		\centering
		\includegraphics[width=\linewidth]{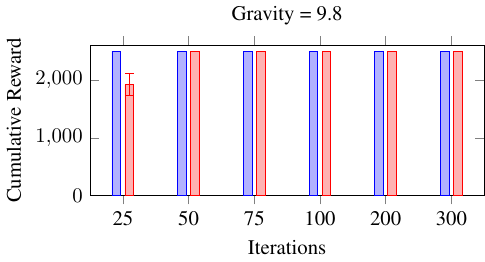}
	\end{subfigure}%
	\begin{subfigure}{.47\textwidth}
		\centering
		\includegraphics[width=\linewidth]{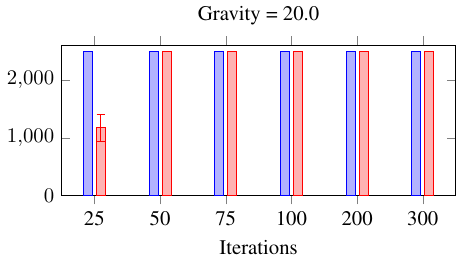}
	\end{subfigure}
	\begin{subfigure}{.5\textwidth}
		\centering
		\includegraphics[width=\linewidth]{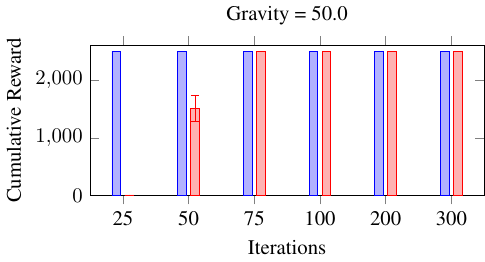}
	\end{subfigure}%
	\begin{subfigure}{.47\textwidth}
		\centering
		\includegraphics[width=\linewidth]{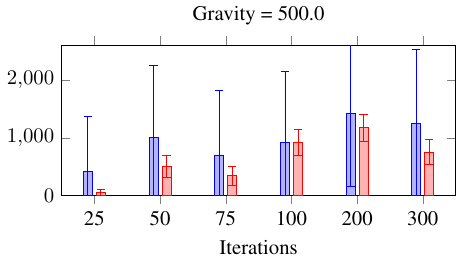}
	\end{subfigure}
	\begin{subfigure}{\textwidth}
		\centering
		\includegraphics[width=0.25\linewidth]{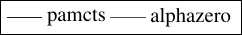}
		\caption*{} 
	\end{subfigure}
	
	\caption{We compare PA-MCTS with $\alpha$-selection with AlphaZero. While both approaches find the optimal policy, PA-MCTS converges significantly faster.}
	\label{figure:az_gravity}
\end{figure}

\textbf{Comparison with AlphaZero:} We use the proposed selection strategy to tune $\alpha$ and compare our results with AlphaZero (presented in \cref{figure:az_gravity}). First, we observe that except for a drastic environmental change ($g=500\,\text{m/s}^2$), both AlphaZero and PA-MCTS (with $\alpha$-selection) converge to the optimal policy. However, we see that PA-MCTS (with $\alpha$-selection) converges significantly faster than AlphaZero; indeed, when $g=50\,\text{m/s}^2$, PA-MCTS takes about a third of the number iterations that AlphaZero takes to converge. In real-world settings, faster convergence is potentially invaluable. \\
\textbf{Runtime Comparison:} We also explore whether re-training the policies by using the updated environmental model is computationally cheaper than executing PA-MCTS. \cref{tab:masspole_time_constraints} shows the calculated average PA-MCTS execution time for different iteration counts (with $\alpha$-selection), and the retraining time for the DDQN network and the Alphazero network. For a fair comparison, we do not retrain DDQN and AlphaZero from scratch and update the existing networks, which is considerably faster than training from scratch, given the updated environment. We show that it is significantly faster to run PA-MCTS than re-training DDQN or AlphaZero.\\
\textbf{Noisy settings:} In practice, it is difficult to update the environmental model to reflect the exact changes in the non-stationary environment. While the generative models can be updated online using updated sensor measurements, such measurements can be noisy, thereby resulting in a noisy model. To simulate such conditions, we repeat the entire pipeline by adding noise in the updated environmental model, e.g., the agent perceives that $g$ = $g_\text{true} \pm \mathcal{N}(0,\sigma)$, (where $\mathcal{N}$ denotes a Gaussian distribution) for different values of $\sigma$. Note that during execution, the agent must operate under $g_\text{true}$, thereby making this situation particularly challenging. We show extensive results for this setting in the appendix but present the key results in \cref{figure:bar_plots}. We see that even under noisy settings, PA-MCTS is able to leverage the old policy and outperforms AlphaZero (it also outperforms standard MCTS and DDQN by a large margin, which we show in the appendix).

\input{runtime}


%% file: results.tex
\begin{table*}[t!]
\small
\centering
\begin{tabular}{lccccc}
\hline
Environment                                                                        & Setting                   & DDQN                         & MCTS                       & AlphaZero                    & PA-MCTS                      \\ \hline
\multirow{4}{*}{\begin{tabular}[c]{@{}l@{}}Cartpole \\ (varying $g$)\end{tabular}} & g = 9.8                   & \textbf{2500.0 $\pm$ 0.0}   & 846.456$\pm$43.228         & 2403.261$\pm$35.946          & \textbf{2500.0$\pm$0.0}      \\
                                                                                   & g = 20                    & \textbf{2500.0$\pm$0.0}     & 918.022$\pm$46.554         & 2278.90$\pm$52.9             & \textbf{2500.0$\pm$0.0}      \\
                                                                                   & g = 50                    & 22.061$\pm$0.729            & 778.511$\pm$44.90          & 1920.261$\pm$78.547          & \textbf{2500.0$\pm$0.0}      \\
                                                                                   & g = 500                   & 7.083$\pm$0.126             & 111.578$\pm$17.705         & 626.178$\pm$80.091           & \textbf{954.656$\pm$90.150}  \\ \hline
\multirow{5}{*}{\begin{tabular}[c]{@{}l@{}}Cartpole \\ (varying $m$)\end{tabular}} & m = 0.1                   & \textbf{2500.0 $\pm$ 0.0}   & 846.456$\pm$43.228         & 2403.261$\pm$35.946          & \textbf{2500.0$\pm$0.0}      \\
                                                                                   & m = 1.0                   & \textbf{500.0$\pm$0.0}      & 892.956$\pm$42.674         & 1340.006$\pm$86.713          & \textbf{2500.0$\pm$0.0}      \\
                                                                                   & m = 1.2                   & 1249.661$\pm$90.587         & 802.905$\pm$30.748         & \textbf{2093.411$\pm$65.519} & 1715.783$\pm$68.101          \\
                                                                                   & m = 1.3                   & 475.35$\pm$67.96            & 904.078$\pm$38.404         & \textbf{1346.433$\pm$86.238} & 904.078$\pm$38.404           \\
                                                                                   & m = 1.5                   & \textbf{2.0$\pm$0.0}        & \textbf{2.0$\pm$0.0}       & \textbf{2.0$\pm$0.0}         & \textbf{2.0$\pm$0.0}         \\ \hline
\multirow{5}{*}{Frozen Lake}                                                       & {[}1.000, 1.000, 1.000{]} & \textbf{1.0 $\pm$ 0.0}      & \textbf{1.0 $\pm$ 0.0}     & \textbf{1.0 $\pm$ 0.0}       & \textbf{1.0$\pm$0.0}         \\
                                                                                   & {[}0.833, 0.083, 0.083{]} & \textbf{0.830 $\pm$ 0.012}  & 0.806 $\pm$ 0.012          & 0.809$\pm$0.013              & \textbf{0.830$\pm$0.012}     \\
                                                                                   & {[}0.633, 0.183, 0.183{]} & 0.522 $\pm$ 0.016           & 0.56 $\pm$ 0.017           & 0.523$\pm$0.017              & \textbf{0.587$\pm$0.016}     \\
                                                                                   & {[}0.433, 0.283, 0.283{]} & 0.26 $\pm$ 0.014            & 0.764 $\pm$ 0.013          & 0.235$\pm$0.014              & \textbf{0.796$\pm$0.013}     \\
                                                                                   & {[}0.333, 0.333, 0.333{]} & 0.12 $\pm$ 0.01             & 0.866 $\pm$ 0.01           & 0.114$\pm$0.011              & \textbf{0.936$\pm$0.009}     \\ \hline
\multirow{4}{*}{Cliff Walking}                                                     & 0.0                       & \textbf{0.88 $\pm$ 0.00}    & 0.869 $\pm$ 0.001          & 0.874 $\pm$ 0.00             & \textbf{0.88 $\pm$ 0.00}     \\
                                                                                   & 0.1                       & 0.291 $\pm$ 0.016           & \textbf{0.7 $\pm$ 0.011}   & 0.27 $\pm$ 0.014             & \textbf{0.7 $\pm$ 0.011}     \\
                                                                                   & 0.2                       & 0.102 $\pm$ 0.011           & \textbf{0.459 $\pm$ 0.015} & 0.081 $\pm$ 0.009            & \textbf{0.459 $\pm$ 0.015}   \\
                                                                                   & 0.3                       & 0.034 $\pm$ 0.006           & \textbf{0.173 $\pm$ 0.012} & 0.012 $\pm$ 0.004            & \textbf{0.173 $\pm$ 0.012}   \\ \hline
\multirow{4}{*}{Lunar Lander}                                                      & 0.0                       & \textbf{256.034$\pm$2.939}  & -76.867$\pm$3.792          & -286.325 $\pm$ 16.759        & \textbf{256.034 $\pm$ 2.939} \\
                                                                                   & 10.0                      & \textbf{241.676$\pm$4.956}  & -79.745$\pm$4.697          & -250.502 $\pm$ 14.967        & \textbf{241.676 $\pm$ 4.955} \\
                                                                                   & 15.0                      & \textbf{192.877$\pm$9.25}   & -94.906$\pm$6.440          & -255.377 $\pm$ 13.234        & \textbf{192.877 $\pm$ 9.25}  \\
                                                                                   & 20.0                      & \textbf{113.431$\pm$14.113} & -119.177$\pm$8.654         & -253.276 $\pm$ 11.086        & \textbf{113.43 $\pm$ 14.113} \\ \hline
\end{tabular}
\caption{Results for all four environments with varying levels of non-stationarity. For each environment, the degree of \textit{change} increases from top to bottom. We observe that PA-MCTS comprehensively outperforms the baseline approaches, including AlphaZero.}
\label{tab:results}
\end{table*}

%% file: runtime.tex
\begin{table}
\begin{center}
\begin{tabular}{@{}ccc@{}}
\toprule
Method             & Iterations & Execution Time (in seconds) \\ \midrule
PA-MCTS               & 5          & 2.57                        \\
PA-MCTS               & 10         & 4.35                        \\
PA-MCTS               & 15         & 3.51                        \\
PA-MCTS               & 25         & 9.60                        \\
PA-MCTS               & 50         & 25.50                       \\
PA-MCTS               & 100        & 48.73                       \\
PA-MCTS               & 200        & 100.0                       \\
PA-MCTS               & 300        & 248.8                       \\
Re-Train DQN       & 200,000 steps           & 430.0                       \\
Re-Train AlphaZero & NA           & 31,800                       \\ \bottomrule
\end{tabular}
\end{center}
\caption{We show the computation time for running PA-MCTS (with $\alpha$-selection and re-training DQN and AlphaZero (by updating the existing networks instead of training from scratch for a fair comparison). Even under these settings, we show that PA-MCTS is significantly faster.}
\label{tab:masspole_time_constraints}
\vspace{-0.5cm}
\end{table}

%% file: relatedWork.tex
\section{Related Work}

Sequential decision-making in non-stationary environments has been studied from several perspectives. \citet{satia1973markovian} and \citet{white1994markov} consider transition matrices constrained within a pre-specified polytope.
One of the earliest works in this domain, by Satia and Lave~\citep{satia1973markovian}, constrains the state transition function among a set of predefined distributions. White and Eldeib~\citep{white1994markov} consider imprecise transition probabilities by representing the transition function by a finite set of linear inequalities. In both formulations, the transition matrix is constrained within a pre-specified polytope. 
However, as pointed out by \citet{iyengar2005robust}, they do not discuss how such polytopes can be constructed. \citet{iyengar2005robust} introduces the idea of robust MDPs using the concept of uncertain priors, where the transition function can change within a set of functions due to uncertainty. 
Later, \citet{lecarpentier2019non} introduce Non-Stationary Markov Decision Processes (NSMDP), which extend robust MDPs by allowing uncertainty in the reward model in addition to the transition function and use a stronger Lipschitz formulation for the set of possible transition and reward functions. \citet{choi2000hidden} introduce hidden-mode MDPs, which consider a formal model for representing changes in the environment, confined with a set of modes. A broader notion of non-stationarity is introduced by \citet{lecarpentier2019non}, who consider that both the reward function and the transition function can change over time and that the rate of change is bounded through Lipschitz continuity. 
Our formulation is inspired by that of \citet{lecarpentier2019non}. The key difference in our formulation is the representation of that change. 

We also point out that decision-making by an agent that is trained on one task (or environmental conditions) and subsequently provided with another task has also been explored in the domains of transfer learning~\citep{minku2019transfer} and lifelong reinforcement learning~\citep{silver2013lifelong}. However, we specifically look at settings where ``learning'' a new policy is not feasible, or even when it is feasible, decisions must be taken while the updated policy is being learned. Finally, we recognize that approaches combining model-based online search with learning methods have been explored---for example, AlphaZero integrates MCTS with a policy iteration framework~\citep{silver2018general}, while the Search with Amortized Value Estimates (SAVE) algorithm combines model-free Q-learning with MCTS~\citep{hamrick2019combining}. To the best of our knowledge, our work is the first that develop such an approach for non-stationary environments. 

%% file: conclusion.tex
\section{Conclusion}\label{sec:conclusion}
We explore sequential decision-making in non-stationary environments, where the decision-maker faces the dilemma of choosing between accurate but obsolete state-action values from a learning-based approach and unbiased but high-variance estimates from an online search. We present a novel approach, \textit{Policy Augmented MCTS} (PA-MCTS), that combines the strengths of reinforcement learning and online planning in non-stationary environments. We present theoretical results characterizing the performance of MCTS, demonstrating that there is a range of values that will work well in practice. We also present bounds on the error accrued by following PA-MCTS as a policy for sequential decision-making. Through extensive experiments on open-source environments, we show that PA-MCTS outperforms the baselines in terms of performance and convergence time.

%% file: appendix.tex
\newpage
\appendix
\clearpage
\onecolumn
\appendixpage



\input{appendix_theory}

\input{appendix_cartpole}
\input{appendix_frozenlake}
\input{appendix_cliffwalking}
\input{appendix_lunarlander}

%% file: appendix_theory.tex
\section{Extended Theoretical Analysis} \label{app:theory}

This appendix contains detailed proofs for each of the theorems, propositions, and corollaries described in Section 4.2 of the main text. 

\textbf{Theorem 1.} If \ $\forall s, a \colon \sum_{s'\in S} \left| P_{t}(s'\mid a, s) - P_{0}(s'\mid a, s) \right| \leq \eta$, $\forall s, a  \colon |r(s, a)| \leq R$, and the discount factor $\gamma < 1$,  
    then $|Q_{0}^{\pi^{*}_{0}}(s,a)-Q_{t}^{\pi^{*}_{t}}(s,a)| \leq \epsilon \  \forall s, a$, where \ $\epsilon = \frac{\gamma \cdot \eta \cdot R}{(1 - \gamma)^2}$ .

\begin{proof}
Our goal is to find an upper bound $\epsilon$ on the difference between the action values before a change in the environment's transition probabilities and after, i.e.:

\begin{equation*}
    \forall s, a: \left| Q'(s, a) - Q(s, a) \right| \leq \epsilon
\end{equation*}

where $Q$ compactly represents the action-value function in the original environment using an optimal policy (i.e., $Q := Q^{\pi^{*}_{0}}_{0}$) and $Q'$ represents the action-value function in an updated environment at time step $t\in \mathcal{T}$ (i.e., $Q' := Q^{\pi^{*}_{t}}_{t}$). 

Let $V_n(s)$ be the value of state $s$ if we terminate the environment after $n$ steps (following the optimal policy until step $n$). 
Similarly, let $Q_n(s, a)$ be the $Q(s, a)$ if we terminate the environment after $n$ steps.

For $n = 1$, $|V_n'(s) - V_n(s)| = 0$ and $|Q_n'(s, a) - Q_n(s, a)| = 0$, since we can assume immediate rewards are the same. 

Let $\kappa_{n-1} = \max_s |V'_{n-1}(s) - V_{n-1}(s)|$ (i.e., $\kappa_{n-1}$ is the upper bound in the difference of the value).

For an arbitrary $n$, we can bound the action-value function $Q$ as follows
\small
\begin{align*}
\left| Q_n'(s, a) - Q_n(s, a) \right| &= \left| \left[ r(s, a) + \gamma \sum_{s'} T'(s, a, s') \cdot V'_{n-1}(s') \right] - \left[ r(s, a) + \gamma \sum_{s'} T(s, a, s') \cdot V_{n-1}(s') \right] \right| \\
&= \gamma \left| \sum_{s'} T'(s, a, s') \cdot V'_{n-1}(s') - T(s, a, s') \cdot V_{n-1}(s') \right| \\ 
&= \gamma \left| \sum_{s'}  T'(s, a, s') \cdot \left( V'_{n-1}(s') - V_{n-1}(s') \right) + T'(s, a, s') \cdot V_{n-1}(s') - T(s, a, s') \cdot V_{n-1}(s') \right| \\ 
&= \gamma \left| \sum_{s'}  T'(s, a, s') \cdot \left( V'_{n-1}(s') - V_{n-1}(s') \right) + V_{n-1}(s') \cdot \left(  T'(s, a, s') - T(s, a, s') \right) \right|  \\ 
&= \gamma \left| \sum_{s'}  T'(s, a, s') \cdot \left( V'_{n-1}(s') - V_{n-1}(s') \right) + \sum_{s'} V_{n-1}(s') \cdot \left(  T'(s, a, s') - T(s, a, s') \right) \right|  \\ 
&\leq \gamma \left| \sum_{s'}  T'(s, a, s') \cdot \left( V'_{n-1}(s') - V_{n-1}(s') \right) \right| + \gamma \left| \sum_{s'} V_{n-1}(s') \cdot \left(  T'(s, a, s') - T(s, a, s') \right) \right|  \\
&\leq \gamma \sum_{s'} \left| T'(s, a, s') \cdot \left( V'_{n-1}(s') - V_{n-1}(s') \right) \right| + \gamma \sum_{s'} \left| V_{n-1}(s') \cdot \left(  T'(s, a, s') - T(s, a, s') \right) \right|  \\
\end{align*}
\begin{align*}
&= \gamma \sum_{s'} \left| T'(s, a, s') \right| \cdot \left| V'_{n-1}(s') - V_{n-1}(s') \right| + \gamma \sum_{s'} \left| V_{n-1}(s') \right| \cdot \left| T'(s, a, s') - T(s, a, s') \right|  \\ 
&\leq \gamma \sum_{s'} T'(s, a, s') \cdot \kappa_{n-1} + \gamma \sum_{s'} \frac{R}{1-\gamma} \cdot \left|  T'(s, a, s') - T(s, a, s') \right|  \\ 
&= \gamma \cdot \kappa_{n-1} \sum_{s'} T'(s, a, s') + \gamma \frac{R}{1-\gamma} \sum_{s'} \left|  T'(s, a, s') - T(s, a, s') \right|  \\ 
&\leq \gamma \cdot \kappa_{n-1} \cdot 1 + \gamma \frac{R}{1-\gamma} \epsilon  \\ 
&= \gamma \left[ \epsilon \frac{R}{1 - \gamma} + \kappa_{n-1} \right]
\end{align*}
\normalsize

We can also bound the value function $V$. First, assume  that $V_n'(s) \geq V_n(s)$, then for an arbitrary $n$, we have
\begin{align*}
\left| V_n'(s) - V_n(s) \right| &= V_n'(s) - V_n(s) \\
&= \max_a Q'_n(s, a) - \max_a Q_n(s, a) \\
&= \max_a \left[ Q'_n(s, a) - Q_n(s, a) + Q_n(s, a) \right] - \max_a Q_n(s, a) \\
&\leq \max_a \left[ \gamma \left[ \epsilon \frac{R}{1 - \gamma} + \kappa_{n-1} \right] + Q_n(s, a) \right] - \max_a Q_n(s, a) \\
&= \gamma \left[ \epsilon \frac{R}{1 - \gamma} + \kappa_{n-1} \right] + \max_a Q_n(s, a) - \max_a Q_n(s, a) \\
&\leq \gamma \left[ \epsilon \frac{R}{1 - \gamma} + \kappa_{n-1} \right]
\end{align*}

Second, assume that $V_n'(s) < V_n(s)$, then for an arbitrary $n$, we have
\begin{align*}
\left| V_n'(s) - V_n(s) \right| &= V_n(s) - V_n'(s) \\
&= \max_a Q_n(s, a) - \max_a Q'_n(s, a)  \\
&=  \max_a Q_n(s, a) - \max_a \left[ Q'_n(s, a) - Q_n(s, a) + Q_n(s, a) \right] \\
&\leq  \max_a Q_n(s, a) - \max_a \left[ -\gamma \left[ \epsilon \frac{R}{1 - \gamma} + \kappa_{n-1} \right] + Q_n(s, a) \right] \\
&=  \max_a Q_n(s, a) +  \gamma \left[ \epsilon \frac{R}{1 - \gamma} + \kappa_{n-1} \right] - \max_a Q_n(s, a)  \\
&\leq \gamma \left[ \epsilon \frac{R}{1 - \gamma} + \kappa_{n-1} \right]
\end{align*}
Thus, $| V_n'(s) - V_n(s)| \leq \gamma \left[ \epsilon \frac{R}{1 - \gamma} + \kappa_{n-1} \right]$ in both cases.

Therefore,
\begin{equation*}
    \kappa_n \leq \gamma \left[ \epsilon \frac{R}{1 - \gamma} + \kappa_{n-1} \right]
\end{equation*}
\begin{equation*}
    \kappa_n \leq \underbrace{\gamma \left( \epsilon \frac{R}{1 - \gamma} + \gamma \left( \epsilon \frac{R}{1 - \gamma} +  \gamma \left( \epsilon \frac{R}{1 - \gamma} +  \ldots \right) \right) \right)}_{n~\text{terms, plus}~\kappa_0 = 0}
\end{equation*}

Hence,
\begin{align*}
\kappa_n &\leq \sum_{i = 1}^n \gamma^i \epsilon \frac{R}{1 - \gamma}  \\
 &= \gamma \cdot \epsilon \frac{R}{1 - \gamma} \sum_{i = 0}^{n - 1} \gamma^i  \\
 &\leq \gamma \cdot \epsilon \frac{R}{1 - \gamma} \sum_{i = 0}^{\infty} \gamma^i  \\
 &= \gamma \cdot \epsilon \frac{R}{1 - \gamma} \cdot \frac{1}{1 - \gamma}  \\
 &= \frac{\gamma \cdot \epsilon \cdot R}{(1 - \gamma)^2}
\end{align*}

\end{proof}


\textbf{Theorem 2.} \textit{If $\alpha\epsilon + (1-\alpha)\delta \leq \frac{\psi_{t}}{2}$, PA-MCTS is guaranteed to select the optimal action at time step~$t$.}

\begin{proof}

PA-MCTS is guaranteed to select the optimal action when the following inequality holds: 
\begin{equation}
    \alpha Q^{\pi^{*}_{0}}_{0}(s, a'_{t}) + (1-\alpha)\overline{G}_{t}(s, a'_{t}) \leq \alpha Q^{\pi^{*}_{0}}_{0}(s, a_{t}^{*}) + (1-\alpha)\overline{G}_{t}(s, a_{t}^{*}) \label{a:p_optimal:main_inequality} 
\end{equation}

Recall that $\delta$ denotes the bound on the error of the values estimated by MCTS when it is stopped, i.e., $|Q_{t}^{\pi^{*}_{t}}(s,a) - \overline{G}_{t}(s, a)|_{\infty} \leq \delta \ \ \ \ \forall  \ \ t \in \mathcal{T}$. Using this definition and Theorem 1, $Q^{\pi^{*}_{t}}_{t}(s, a)$ can be bounded with respect to the estimates $Q^{\pi^{*}_{0}}_{0}(s, a)$ and $\overline{G}_{t}(s, a)$:
\begin{equation}
\begin{aligned}
    & Q^{\pi^{*}_{0}}_{0}(s, a) - \epsilon \leq Q^{\pi^{*}_{t}}_{t}(s, a) \leq Q^{\pi^{*}_{0}}_{0}(s, a) + \epsilon \\
    & \overline{G}_{t}(s, a) - \delta \leq Q^{\pi^{*}_{t}}_{t}(s, a) \leq \overline{G}_{t}(s, a) + \delta \label{a:eq:p1:approxbound_d}
\end{aligned}
\end{equation}

By substituting $ Q^{\pi^{*}_{t}}_{t}(s, a)$ for $Q^{\pi^{*}_{0}}_{0}$ and $\overline{G}_{t}$ in \cref{a:p_optimal:main_inequality} by using inequalities in~\cref{a:eq:p1:approxbound_d} and rearranging, we can find the conditions under which PA-MCTS chooses the optimal action (recall that $\psi_{t}(s) := Q^{\pi^{*}_{t}}_{t}(s,a_{t}^*) - Q^{\pi^{*}_{t}}_{t}(s,a_{t}')$):

\begin{align}
    \alpha Q^{\pi^{*}_{0}}_{0}(s, a'_{t}) + (1-\alpha)\overline{G}_{t}(s, a'_{t}) &\leq \alpha Q^{\pi^{*}_{0}}_{0}(s, a_{t}^{*}) + (1-\alpha)\overline{G}_{t}(s, a_{t}^{*}) \nonumber \\
    \Rightarrow \alpha(Q^{\pi^{*}_{t}}_{t}(s, a'_{t}) + \epsilon) + (1-\alpha)(Q^{\pi^{*}_{t}}_{t}(s, a'_{t}) + \delta) &\leq \alpha(Q^{\pi^{*}_{t}}_{t}(s, a^{*}_{t}) - \epsilon) + (1-\alpha)(Q^{\pi^{*}_{t}}_{t}(s, a^{*}_{t}) - \delta) \nonumber \\
    \Rightarrow 2(\alpha\epsilon + (1-\alpha)\delta) &\leq Q^{\pi^{*}_{t}}_{t}(s, a^{*}_{t}) - Q^{\pi^{*}_{t}}_{t}(s, a'_{t}) \nonumber \\ 
    \Rightarrow 2(\alpha\epsilon + (1-\alpha)\delta) &\leq \psi_t \nonumber \\
    \Rightarrow \alpha\epsilon + (1-\alpha)\delta &\leq \frac{\psi_t}{2} \nonumber
\end{align}
\end{proof}


\textbf{Corollary 2.1.} \textit{$\psi_{t} \leq \psi_{0} + 2\epsilon$}

\begin{proof}

It is important to note that the optimal and second best actions with respect to $Q^{\pi^{*}_{0}}_{0}$ and $Q^{\pi^{*}_{t}}_{t}$ may not be the same, i.e., it is possible that $a_{0}^* \neq a_{t}^*$ and $a_{0}' \neq a_{t}'$. There exist the following cases depending on the relationship between $a_{0}'$ and  $a_{t}'$: 

\begin{itemize}
    \item If $a'_{0} = a'_{t}$, we first upper bound $Q^{\pi^{*}_{t}}_{t}(s,a_{t}^*)$. We know that for any action $a\in \mathcal{A}$ 
    \begin{equation}
    Q^{\pi^{*}_{t}}_{t}(s, a) \leq Q^{\pi^{*}_{0}}_{0}(s, a_{0}^{*}) + \epsilon \nonumber
    \end{equation}
    Therefore, 
    \begin{equation}
        Q^{\pi^{*}_{t}}_{t}(s, a_{t}^{*}) \leq Q^{\pi^{*}_{0}}_{0}(s, a_{0}^{*}) + \epsilon \label{a:c1:a*}
    \end{equation}
    We can then simply use the definition of $\epsilon$ to lower bound $Q^{\pi^{*}_{t}}_{t}(s, a'_{t})$ :
    \begin{equation}
        Q^{\pi^{*}_{t}}_{t}(s, a'_{t}) \ge Q^{\pi^{*}_{0}}_{0}(s, a'_{0}) - \epsilon \label{a:c1:a'}
    \end{equation}
    Finally, by combining \cref{a:c1:a*} and \cref{a:c1:a'}, we can upper bound $\psi_{t}$ 
    \begin{align}
        Q^{\pi^{*}_{t}}_{t}(s,a_{t}^*) - Q^{\pi^{*}_{t}}_{t}(s,a_{t}') &\leq (Q^{\pi^{*}_{0}}_{0}(s, a_{0}^{*}) + \epsilon) - (Q^{\pi^{*}_{0}}_{0}(s, a'_{0}) - \epsilon) \nonumber \\
        \Rightarrow Q^{\pi^{*}_{t}}_{t}(s,a_{t}^*) - Q^{\pi^{*}_{t}}_{t}(s,a_{t}') &\leq Q^{\pi^{*}_{0}}_{0}(s, a_{0}^{*}) - Q^{\pi^{*}_{0}}_{0}(s, a'_{0}) + 2\epsilon \nonumber \\
        \Rightarrow \psi_{t} &\leq \psi_{0} + 2\epsilon \nonumber
    \end{align}
    
    \item If $a'_{0} = a^{*}_{t}$, i.e., the second best action at time step 0 became the optimal action at time step~$t$, then this implies that $a^{*}_{0} \neq a^{*}_{t}$. This implies: 
    \begin{equation}
        \psi_{t} \leq 2\epsilon \nonumber
    \end{equation}
    as it would otherwise be impossible for the optimal action at timestep 0 to become sub-optimal at timestep $t$. Since $\psi_{0}$ is positive by definition, this implies that $\psi_{t} \leq \psi_{0} + 2\epsilon$
    
    \item If $a'_{0} \notin \{a^{*}_{t}, a'_{t}\}$, then it follows from the definition of $a_{t}'$ that $Q^{\pi^{*}_{0}}_{0}(s, a'_{t}) \leq Q^{\pi^{*}_{0}}_{0}(s,a'_{0})$ and $Q^{\pi^{*}_{t}}_{t}(s, a'_{t}) \ge Q^{\pi^{*}_{t}}_{t}(s, a'_{0})$. 
    
    Let us assume that $Q^{\pi^{*}_{t}}_{t}(s, a'_{t}) < Q^{\pi^{*}_{0}}_{0}(s,a'_{0}) - \epsilon$. Then, by substituting $Q^{\pi^{*}_{0}}_{0}(s,a'_{0}) - \epsilon$ for $Q^{\pi^{*}_{t}}_{t}(s, a'_{0})$ using \cref{a:eq:p1:approxbound_d}, this implies that $Q^{\pi^{*}_{t}}_{t}(s, a'_{t}) < Q^{\pi^{*}_{t}}_{t}(s, a'_{0})$. This is a contradiction given the above inequalities. Therefore, through proof by contradiction, $Q^{\pi^{*}_{t}}_{t}(s, a'_{t}) \ge Q^{\pi^{*}_{0}}_{0}(s,a'_{0}) - \epsilon$. Similar to the $a'_{0} = a'_{t}$ case above, we can combine this with the upper bound in \cref{a:c1:a*} to find that 
    \begin{equation}
        \psi_{t} \leq 2\epsilon \nonumber
    \end{equation}
\end{itemize}

In each case, $\psi_{t} \leq 2\epsilon$. \end{proof}








\textbf{Corollary 2.2.} \textit{If $\alpha\epsilon + (1-\alpha)\delta \leq \frac{\psi_{0}}{2} + \epsilon$, PA-MCTS is guaranteed to select the optimal action at decision epoch $t$.}

\begin{proof}
    Using Corollary 2.1, we substitute $\psi_{t}$ in terms of $\psi_{0}$ in Theorem 2. 
    \begin{align}
        \alpha\epsilon + (1-\alpha)\delta &\leq \frac{\psi_{t}}{2} \nonumber \\
        \Rightarrow \alpha\epsilon + (1-\alpha)\delta &\leq \frac{\psi_{0} + 2\epsilon}{2} \nonumber \\
        \Rightarrow \alpha\epsilon + (1-\alpha)\delta &\leq \frac{\psi_{0}}{2} + \epsilon \nonumber
    \end{align}\end{proof}


\textbf{Corollary 2.3.} \textit{PA-MCTS will choose the optimal one-step action if}
\begin{equation}
    \begin{cases}
        \frac{-\delta}{\epsilon - \delta} \leq \alpha \leq \frac{\psi_{0}}{2(\epsilon - \delta)} + 1 &\text{if $\epsilon > \delta$}\\
        \frac{-\delta}{\epsilon - \delta} \ge \alpha \ge \frac{\psi_{0}}{2(\epsilon - \delta)} + 1 &\text{if $\epsilon < \delta$.}
    \end{cases} \nonumber
\end{equation} 

\begin{proof}
    Using $0 \le \alpha \le 1$, Corollary 2.2 can be rearranged to solve for $\alpha$:
    
    \begin{align}
        &0 \leq \alpha\epsilon + (1-\alpha)\delta \leq \frac{\psi_{0}}{2} + \epsilon \nonumber \\
        \Rightarrow \ &0 \leq \alpha\epsilon - \alpha\delta + \delta \leq \frac{\psi_{0}}{2} + \epsilon \nonumber \\
        \Rightarrow \ &-\delta \leq \alpha\epsilon - \alpha\delta \leq \frac{\psi_{0}}{2} + \epsilon - \delta \nonumber \\
        \Rightarrow \ &\begin{cases}
            \frac{-\delta}{\epsilon - \delta} \leq \alpha \leq \frac{\psi_{0}}{2(\epsilon - \delta)} + 1 &\text{if $\epsilon > \delta$}\\
            \frac{-\delta}{\epsilon - \delta} \ge \alpha \ge \frac{\psi_{0}}{2(\epsilon - \delta)} + 1 &\text{if $\epsilon < \delta$.}
        \end{cases} \nonumber
    \end{align}
\end{proof}







\textbf{Proposition 1.} \textit{If PA-MCTS and MCTS choose different actions, PA-MCTS's chosen action will have a higher $Q^{\pi^{*}_{t}}_{t}$ value than MCTS if $2\epsilon\ \leq \zeta^{m}_{t}$, where\footnote{The superscript $m$ in $\zeta^{m}_{t}$ represents the bounds with respect to MCTS and is not an index.} $\zeta^{m}_{t} := (Q^{\pi^{*}_{0}}_{0}(s, \tilde{a}) + \overline{G}_{t}(s,  \tilde{a})) - (Q^{\pi^{*}_{0}}_{0}(s, a_m) + \overline{G}_{t}(s,  a_m))$, $\tilde{a} := \argmax_{a \in \mathcal{A}_s} \ \ \ \alpha Q^{\pi^{*}_{0}}_{0}(s, a) + (1-\alpha)\overline{G}_{t}(s,  a)$, and $a_m := \argmax_{a \in \mathcal{A}_s} \overline{G}_{t}(s,  a)$. }

\begin{proof}
    PA-MCTS's chosen action $\tilde{a}$ has a higher $Q^{\pi^{*}_{t}}_{t}$ value than MCTS's chosen action $a_m$ when
\begin{equation}
    0 \le Q^{\pi^{*}_{t}}_{t}(s,\tilde{a}) - Q^{\pi^{*}_{t}}_{t}(s,a_m) \label{a:eq:p2:0inequality}
\end{equation}
If PA-MCTS and MCTS choose different actions, then it must be that $\overline{G}_{t}(s, a_m) \ge  \overline{G}_{t}(s,  \tilde{a})$ and $\alpha Q^{\pi^{*}_{0}}_{0}(s, a_m) + (1-\alpha)\overline{G}_{t}(s,  a_m) \leq \alpha Q^{\pi^{*}_{0}}_{0}(s, \tilde{a}) + (1-\alpha)\overline{G}_{t}(s, \tilde{a})$.
Then, we can bound $Q^{\pi^{*}_{t}}_{t}$ with respect to $Q^{\pi^{*}_{0}}_{0}$ and $\epsilon$:
\begin{equation}
    Q^{\pi^{*}_{0}}_{0}(s, \tilde{a}) - Q^{\pi^{*}_{0}}_{0}(s, a_m) - 2\epsilon \leq Q^{\pi^{*}_{t}}_{t}(s,\tilde{a}) - Q^{\pi^{*}_{t}}_{t}(s,a_m) \nonumber
\end{equation}
Next, we rewrite this bound in terms of $\zeta^{m}_{t}$ and $\epsilon$:
\begin{align}
    Q^{\pi^{*}_{0}}_{0}(s, \tilde{a}) - Q^{\pi^{*}_{0}}_{0}(s, a_m) - 2\epsilon &\leq Q^{\pi^{*}_{t}}_{t}(s,\tilde{a}) - Q^{\pi^{*}_{t}}_{t}(s,a_m) \nonumber\\ 
    \Rightarrow Q^{\pi^{*}_{0}}_{0}(s, \tilde{a}) - Q^{\pi^{*}_{0}}_{0}(s, a_m) - 2\epsilon + \overline{G}_{t}(s,  a_m) - \overline{G}_{t}(s,  a_m) &\leq Q^{\pi^{*}_{t}}_{t}(s,\tilde{a}) - Q^{\pi^{*}_{t}}_{t}(s,a_m) \label{a:p1:2}\\ 
    \Rightarrow Q^{\pi^{*}_{0}}_{0}(s, \tilde{a}) - Q^{\pi^{*}_{0}}_{0}(s, a_m) - 2\epsilon + \overline{G}_{t}(s,  \tilde{a}) - \overline{G}_{t}(s,  a_m) &\leq Q^{\pi^{*}_{t}}_{t}(s,\tilde{a}) - Q^{\pi^{*}_{t}}_{t}(s,a_m) \label{a:p1:3}\\ 
    \Rightarrow (Q^{\pi^{*}_{0}}_{0}(s, \tilde{a}) + \overline{G}_{t}(s,  \tilde{a})) - (Q^{\pi^{*}_{0}}_{0}(s, a_m) + \overline{G}_{t}(s,  a_m)) - 2\epsilon &\leq Q^{\pi^{*}_{t}}_{t}(s,\tilde{a}) - Q^{\pi^{*}_{t}}_{t}(s,a_m) \label{a:p1:4} \\ 
    \Rightarrow \zeta^{m}_{t} - 2\epsilon &\leq Q^{\pi^{*}_{t}}_{t}(s,\tilde{a}) - Q^{\pi^{*}_{t}}_{t}(s,a_m) \nonumber
\end{align}
where in \cref{a:p1:2} we add and subtract $\overline{G}_{t}(s,  a_m)$ to the left hand side, in \cref{a:p1:3} we use $\overline{G}_{t}(s, a_m) \ge  \overline{G}_{t}(s,  \tilde{a})$ to substitute $\overline{G}_{t}(s,  \tilde{a})$ for $\overline{G}_{t}(s, a_m)$, and in \cref{a:p1:4} we rearrange the left hand side. 

Therefore, it follows from \cref{a:eq:p2:0inequality} that PA-MCTS chooses an action with higher $Q^{\pi^{*}_{t}}_{t}$ value when $0 < \zeta^{m}_{t} - 2\epsilon$, i.e., $2\epsilon\ \leq \zeta^{m}_{t}$.
\end{proof}

\textbf{Proposition 2.}
\textit{If the action $\tilde{a}$ chosen by PA-MCTS is different than $a_g := \argmax_{a \in \mathcal{A}_s} Q^{\pi^{*}_{0}}_{0}(s,a)$ at state $s$, then PA-MCTS's chosen action $\tilde{a}$ will have a higher $Q^{\pi^{*}_{t}}_{t}$ value than action $a_g$ if $2\delta\ \leq \zeta^{q}_{t}$ where\footnote{Again, the superscript $q$ only denotes bounds with respect to the $Q$-values and is not an index.} $\zeta^{q}_{t} := (Q^{\pi^{*}_{0}}_{0}(s, \tilde{a}) + \overline{G}_{t}(s, \tilde{a})) - (Q^{\pi^{*}_{0}}_{0}(s, a_g) + \overline{G}_{t}(s,  a_g))$, $\tilde{a} := \argmax_{a \in \mathcal{A}_s} \alpha Q^{\pi^{*}_{0}}_{0}(s, a) + (1-\alpha)\overline{G}_{t}(s, a)$, and $a_g := \argmax_{a \in \mathcal{A}_s} Q^{\pi^{*}_{0}}_{0}(s,  a)$.}

\begin{proof}
    PA-MCTS's chosen action $\tilde{a}$ has a higher $Q^{\pi^{*}_{t}}_{t}$ value than the action selected through $Q^{\pi^{*}_{0}}_{0}$-values $a_g$ when
    \begin{equation}
        0 \le Q^{\pi^{*}_{t}}_{t}(s,\tilde{a}) - Q^{\pi^{*}_{t}}_{t}(s,a_g) \label{a:eq:p3:0inequality}
    \end{equation}
    
    If PA-MCTS and action selection through $Q^{\pi^{*}_{0}}_{0}$-values choose different actions, then it must be that $Q^{\pi^{*}_{0}}_{0}(s, a_g) \ge  Q^{\pi^{*}_{0}}_{0}(s,  \tilde{a})$ and $\alpha Q^{\pi^{*}_{0}}_{0}(s, a_g) + (1-\alpha)\overline{G}_{t}(s,  a_g) \leq \alpha Q^{\pi^{*}_{0}}_{0}(s, \tilde{a}) + (1-\alpha)\overline{G}_{t}(s, \tilde{a})$.
    Then, we can bound $Q^{\pi^{*}_{t}}_{t}$ with respect to $\overline{G}_{t}$ and $\delta$:
    \begin{equation}
        \overline{G}_{t}(s, \tilde{a}) - \overline{G}_{t}(s,  a_g) - 2\delta \leq Q^{\pi^{*}_{t}}_{t}(s,\tilde{a}) - Q^{\pi^{*}_{t}}_{t}(s,a_g)\nonumber
    \end{equation}
    Next, we can rewrite this bound in terms of $\zeta^{q}_{t}$ and $\delta$:
    
    \begin{align}
        \overline{G}_{t}(s, \tilde{a}) - \overline{G}_{t}(s,  a_g) - 2\delta &\leq Q^{\pi^{*}_{t}}_{t}(s,\tilde{a}) - Q^{\pi^{*}_{t}}_{t}(s,a_g)\nonumber\\ 
        \Rightarrow \overline{G}_{t}(s, \tilde{a}) - \overline{G}_{t}(s,  a_g) - 2\delta + Q^{\pi^{*}_{0}}_{0}(s,  a_g) - Q^{\pi^{*}_{0}}_{0}(s,  a_g) &\leq Q^{\pi^{*}_{t}}_{t}(s,\tilde{a}) - Q^{\pi^{*}_{t}}_{t}(s,a_g) \label{a:p2:2}\\ 
        \Rightarrow \overline{G}_{t}(s, \tilde{a}) - \overline{G}_{t}(s,  a_g) - 2\delta + Q^{\pi^{*}_{0}}_{0}(s,  \tilde{a}) - Q^{\pi^{*}_{0}}_{0}(s,  a_g) &\leq Q^{\pi^{*}_{t}}_{t}(s,\tilde{a}) - Q^{\pi^{*}_{t}}_{t}(s,a_g) \label{a:p2:3}\\ 
        \Rightarrow (Q^{\pi^{*}_{0}}_{0}(s, \tilde{a}) + \overline{G}_{t}(s,  \tilde{a})) - (Q^{\pi^{*}_{0}}_{0}(s, a_g) + \overline{G}_{t}(s,  a_g)) - 2\delta &\leq Q^{\pi^{*}_{t}}_{t}(s,\tilde{a}) - Q^{\pi^{*}_{t}}_{t}(s,a_g) \label{a:p2:4} \\ 
        \Rightarrow \zeta^{g}_{t} - 2\delta &\leq Q^{\pi^{*}_{t}}_{t}(s,\tilde{a}) - Q^{\pi^{*}_{t}}_{t}(s,a_g) \nonumber
    \end{align}
    
    where in \cref{a:p2:2} we introduce $Q^{\pi^{*}_{0}}_{0}(s, a_g)$ by adding zero to the left hand side, in \cref{a:p2:3} we use $Q^{\pi^{*}_{0}}_{0}(s, a_g) \ge  Q^{\pi^{*}_{0}}_{0}(s,  \tilde{a})$ to substitute $Q^{\pi^{*}_{0}}_{0}(s, \tilde{a})$ for $Q^{\pi^{*}_{0}}_{0}(s, a_g)$, and in \cref{a:p2:4} we rearrange the left hand side. 

Therefore, it follows from \cref{a:eq:p3:0inequality} that PA-MCTS chooses an action with higher $Q^{\pi^{*}_{t}}_{t}$ value when $0 < \zeta^{g}_{t} - 2\delta$, i.e., $2\delta\ \leq \zeta^{g}_{t}$.
\end{proof}

\textbf{Theorem 3.} \textit{When PA-MCTS is used for sequential decision making, the maximum difference between the return from an optimal policy and the return from following PA-MCTS is at most $\frac{2(\alpha\epsilon - \alpha\delta + \delta)}{1-\gamma}$.}

\begin{proof}
    The proof closely follows prior work by \citet{wray2019abstractions}.
    Let $\mathcal{A}'_{s}$ denote the set of actions that can be taken by PA-MCTS in state $s$ at any decision epoch. Given $\epsilon$, $\delta$, and $a \in \mathcal{A}'_{s}$, it must be that 
    \begin{equation}
        \alpha Q^{\pi^{*}_{0}}_{0}(s,a) + (1-\alpha)\overline{G}_{t}(s, a) \ge \alpha Q^{\pi^{*}_{0}}_{0}(s,a^{*}_{t}) + (1-\alpha)\overline{G}_{t}(s, a^{*}_{t}) \label{a:t2:a'}
    \end{equation}
    
    for $a$ to be chosen by PA-MCTS. We can substitute $ Q^{\pi^{*}_{t}}_{t}(s, a)$ for $Q^{\pi^{*}_{0}}_{0}$ and $\overline{G}_{t}$ in \cref{a:t2:a'} to get
    \begin{align}
        \alpha Q^{\pi^{*}_{0}}_{0}(s, a) + (1-\alpha)\overline{G}_{t}(s, a) &\ge \alpha Q^{\pi^{*}_{0}}_{0}(s, a_{t}^{*}) + (1-\alpha)\overline{G}_{t}(s, a_{t}^{*}) \nonumber \\
        \Rightarrow \alpha(Q^{\pi^{*}_{t}}_{t}(s, a) + \epsilon) + (1-\alpha)(Q^{\pi^{*}_{t}}_{t}(s, a) + \delta) &\ge \alpha(Q^{\pi^{*}_{t}}_{t}(s, a^{*}_{t}) - \epsilon) + (1-\alpha)(Q^{\pi^{*}_{t}}_{t}(s, a^{*}_{t}) - \delta) \nonumber \\
        \Rightarrow 2(\alpha\epsilon + (1-\alpha)\delta) &\ge Q^{\pi^{*}_{t}}_{t}(s, a^{*}_{t}) - Q^{\pi^{*}_{t}}_{t}(s, a) \nonumber \\ 
        \Rightarrow 2(\alpha\epsilon + (1-\alpha)\delta) &\ge V^{\pi^{*}_{t}}_{t}(s) - Q^{\pi^{*}_{t}}_{t}(s, a) \nonumber
    \end{align}
    where 
    \begin{equation}
        V^{\pi}_{t}(s) = R_{t}(s,\pi(s)) + \gamma \sum_{s'\in \mathcal{S}} P_{t}(s'| s, a)V^{\pi}(s') \label{a:bellman}
    \end{equation}
    where $R_{t}(s, a)$ is the reward for taking action $a$ at $s$ at time step $t$, $\pi(s)$ is policy $\pi$'s distribution over actions given $s$, $\gamma$ is discount factor, and $P_{t}(s'|s,a)$ is the probability of reaching state $s'$ when taking $a$ at $s$ at time step $t$. From \cref{a:bellman}, the optimal value function can be expressed as $V^{\pi^{*}_{t}}_{t}(s) = \argmax_a Q^{\pi^{*}_{t}}_{t}(s,a)$.
    
    Therefore
    \begin{equation}
        \mathcal{A}'_{s} = \{a \in \mathcal{A}_{s} \,|\, V^{\pi^{*}_{t}}_{t}(s) - Q^{\pi^{*}_{t}}_{t}(s, a) \leq 2(\alpha\epsilon + (1-\alpha)\delta) \} \nonumber
    \end{equation}
    
    Now, let $V^{\tilde{\pi}, k}_{t}(s)$ be $k$ applications of \cref{a:bellman} following PA-MCTS's policy $\tilde{\pi}$, i.e.,  $\tilde{\pi} \in \Pi'$ where $\Pi' = \{\pi|\pi(s) \in \mathcal{A}'_{s}, \forall s\in \mathcal{S}\}$. Also let $Q_{t}^{\pi^{*}_{t}, k}(s,\tilde{\pi}(s))$ be a one-step action deviation following $\tilde{\pi}$, after which an optimal policy $\pi^{*}_{t}$ is followed for the remaining $k-1$ iterations. 
    
    Next, we show that for any state $s\in \mathcal{S}$:
    \begin{equation}
        V^{\tilde{\pi}, k}_{t}(s) \ge V^{\pi^{*}_{t}, k}_{t}(s) - \sum_{\tau=0}^{k} \gamma^{\tau} 2(\alpha\epsilon + (1-\alpha)\delta) \label{a:t2:induction_prove}
    \end{equation}
    by induction on $k$ iterations of \cref{a:bellman}. 
    
    \textbf{Base Case}: At $k=0$, by the definition of $\tilde{\pi}(s) \in \mathcal{A}'_{s}$, we have: 
    \begin{align}
        & \ V^{\pi^{*}_{t}, 0}_{t}(s) - Q_{t}^{\pi^{*}_{t}, 0}(s,\tilde{\pi}(s)) \leq 2(\alpha\epsilon + (1-\alpha)\delta) \nonumber\\
        \Rightarrow & \ V^{\pi^{*}_{t}, 0}_{t}(s) - R_{t}(s, \tilde{\pi}(s)) \leq 2(\alpha\epsilon + (1-\alpha)\delta) \nonumber\\
        \Rightarrow & \ V^{\pi^{*}_{t}, 0}_{t}(s) - V^{\tilde{\pi}, 0}_{t}(s) \leq 2(\alpha\epsilon + (1-\alpha)\delta) \nonumber\\
        \Rightarrow & \ V^{\tilde{\pi}, 0}_{t}(s) \ge V^{\pi^{*}_{t}, 0}_{t}(s) - 2(\alpha\epsilon + (1-\alpha)\delta) \nonumber
    \end{align}
    
    \textbf{Induction Step}: Assume for $k-1$ the induction hypothesis: 
    \begin{equation}
        V^{\tilde{\pi}, k-1}_{t}(s) \ge V^{\pi^{*}_{t}, k-1}_{t}(s) - \sum_{\tau=0}^{k-1} \gamma^{\tau} 2(\alpha\epsilon + (1-\alpha)\delta) \nonumber
    \end{equation}
    is true. We now show that \cref{a:t2:induction_prove} is true for $k$:
    
    \begin{align}
        & \ V^{\tilde{\pi}, k}_{t}(s) = R_{t}(s, \tilde{\pi}(s)) + \gamma \sum_{s'\in \mathcal{S}} P_{t}(s'| s, a)V^{\tilde{\pi}, k-1}(s') \label{a:t2:induction1}\\
        \Rightarrow & \ V^{\tilde{\pi}, k}_{t}(s) \ge R_{t}(s, \tilde{\pi}(s)) + \gamma \sum_{s'\in \mathcal{S}} P_{t}(s'| s, a) (V^{\pi^{*}_{t}, k-1}(s')-\sum_{\tau=0}^{k-1}\gamma^{\tau}2(\alpha\epsilon + (1-\alpha)\delta) ) \label{a:t2:induction2} \\
        \Rightarrow & \ V^{\tilde{\pi}, k}_{t}(s) \ge (R_{t}(s, \tilde{\pi}(s)) + \gamma \sum_{s'\in \mathcal{S}} P_{t}(s'| s, a) V^{\pi^{*}_{t}, k-1}(s'))-\sum_{\tau=0}^{k-1}\gamma^{\tau+1}2(\alpha\epsilon + (1-\alpha)\delta) ) \label{a:t2:induction3} \\
        \Rightarrow & \ V^{\tilde{\pi}, k}_{t}(s) \ge Q_{t}^{\pi^{*}_{t}, k}(s,\tilde{\pi}(s)) - \sum_{\tau=0}^{k-1}\gamma^{\tau+1}2(\alpha\epsilon + (1-\alpha)\delta) ) \label{a:t2:induction4} \\
        \Rightarrow & \ V^{\tilde{\pi}, k}_{t}(s) \ge Q_{t}^{\pi^{*}_{t}, k}(s,\tilde{\pi}(s)) - \sum_{\tau=1}^{k}\gamma^{\tau}2(\alpha\epsilon + (1-\alpha)\delta) ) \nonumber
    \end{align}
    where \cref{a:t2:induction1} is by \cref{a:bellman}, \cref{a:t2:induction2} is by the induction hypothesis, \cref{a:t2:induction3} is by rewriting and normalizing, and \cref{a:t2:induction4} is by the definition of $Q_{t}^{\pi^{*}_{t}, k}$. 
    
    By the definition of $\mathcal{A}'_{s}$, we have $V^{\pi^{*}_{t}, k}_{t}(s) - Q_{t}^{\pi^{*}_{t}, k}(s, \tilde{\pi}(s) \leq 2(\alpha\epsilon + (1-\alpha)\delta)$. By rearranging this, applying to $Q_{t}^{\pi^{*}_{t}, k}$ and grouping $2(\alpha\epsilon + (1-\alpha)\delta)$ in the sum, we find
    \begin{align}
        V^{\tilde{\pi}, k}_{t}(s) &\ge V^{\pi^{*}_{t}, k}_{t}(s) - 2(\alpha\epsilon + (1-\alpha)\delta) - \sum_{\tau=1}^{t} \gamma^{\tau} 2(\alpha\epsilon + (1-\alpha)\delta) \nonumber \\
        &= V^{\pi^{*}_{t}, k}_{t}(s) - \sum_{\tau=0}^{t} \gamma^{\tau} 2(\alpha\epsilon + (1-\alpha)\delta) \nonumber
    \end{align}
    
    Thus, by induction on $k$, we have shown that \cref{a:t2:induction_prove} is true for all $k$. Finally let $k\to\infty$ and evaluate \cref{a:t2:induction_prove}:
    
    \begin{align}
        V^{\tilde{\pi}, k}_{t}(s) &\ge V^{\pi^{*}_{t}, k}_{t}(s) - \sum_{\tau=0}^{t} \gamma^{\tau} 2(\alpha\epsilon + (1-\alpha)\delta) \label{a:t2:toinfity_1} \\
        &\ge V^{\pi^{*}_{t}, k}_{t}(s) - \sum_{\tau=0}^{\infty} \gamma^{\tau} 2(\alpha\epsilon + (1-\alpha)\delta) \label{a:t2:toinfity_2} \\
        &\ge V^{\pi^{*}_{t}, k}_{t}(s) - \frac{2(\alpha\epsilon + (1-\alpha)\delta)}{1-\gamma} \label{a:t2:toinfity_3} \\
        V^{\pi^{*}_{t}, k}_{t}(s) - V^{\tilde{\pi}, k}_{t}(s) &\leq \frac{2(\alpha\epsilon + (1-\alpha)\delta)}{1-\gamma} \label{a:t2:toinfity_4}
    \end{align}
    where \cref{a:t2:toinfity_1} is by \cref{a:t2:induction_prove}, \cref{a:t2:toinfity_2} is by $\gamma\in[0, 1)$ and $2(\alpha\epsilon + (1-\alpha)\delta)\ge 0$, \cref{a:t2:toinfity_3} is by geometric series, and \cref{a:t2:toinfity_4} is by rearranging. 
    
\end{proof}

%% file: appendix_cartpole.tex
\section{Experimental Results on the CartPole Environment} \label{app:cartpole}
In the main text, we present results on decision-making in non-stationary settings by varying the gravity parameter in the cartpole environment. To gather more robust results, we also vary the total mass of the cartpole system by increasing the mass of the pole.
We fix gravity=9.8$\text{m/s}^2$ and the mass of the \textit{cart} as 1kg, but we vary the mass of the \textit{pole} among \{0.1kg, 1.0kg, 1.2kg, 1.3kg, 1.5kg\}; note that as the mass of the pole increases, the total mass of the system under consideration (i.e., the mass of the cart and the pole) also increases.
We keep all other hyper-parameters the same as the experiments where gravity is changed (shown in the main text).

\subsection{Stationary Environment}
The results for the stationary Environment are the same as the stationary environment for the gravity experiments. We present the results in the main text (Fig. 2 in the main text). We compute each result by averaging across 30 samples.

\subsection{Non-Stationary Environment}


\begin{figure}[!htb]
	\centering
	\begin{subfigure}{.5\textwidth}
		\centering
		\includegraphics[width=\linewidth]{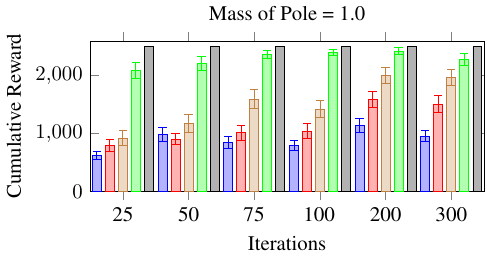}
	\end{subfigure}%
	\begin{subfigure}{.47\textwidth}
		\centering
		\includegraphics[width=\linewidth]{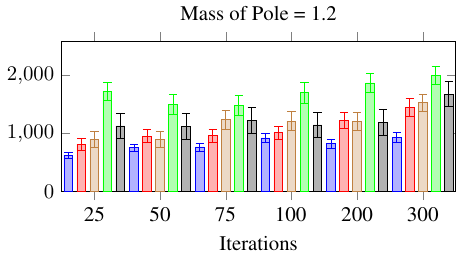}
	\end{subfigure}
	\begin{subfigure}{.5\textwidth}
		\centering
		\includegraphics[width=\linewidth]{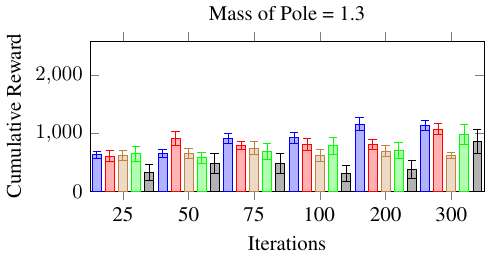}
	\end{subfigure}%
	\begin{subfigure}{.47\textwidth}
		\centering
		\includegraphics[width=\linewidth]{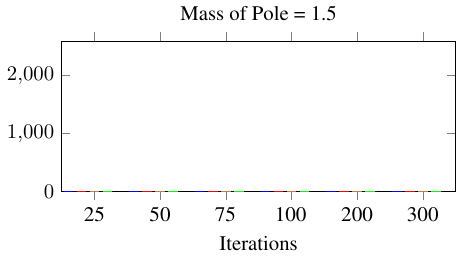}
	\end{subfigure}
	\begin{subfigure}{\textwidth}
		\centering
		\includegraphics[width=0.8\linewidth]{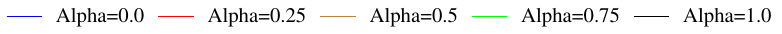}
		\caption*{} 
	\end{subfigure}
	
	\caption{We show the average cumulative reward obtained by PA-MCTS, standard MCTS ($\alpha=0)$, and DDQN ($\alpha=1$) for different environmental changes. The mass of the pole changes among \{0.1kg, 1.0kg, 1.2kg, 1.3kg, 1.5kg\} }
	\label{fig:cartpole_masspole}
\end{figure}

We represent the non-stationary environment here by changing the mass of the pole on the cart. We show the results in \cref{fig:cartpole_masspole}. We vary the mass of the pole among $\{$0.1kg, 1.0kg, 1.2kg, 1.3kg, 1.5kg$\}$. Each $\alpha$ value used for PA-MCTS is represented by a different color. For each setting, we vary MCTS iterations among $\{$ 25, 50, 75, 100, 200, 300 $\}$ on 30 samples (test episodes). 
As \cref{fig:cartpole_masspole} shows, DDQN achieves maximum reward in the stationary setting but its performance deteriorates under non-stationarity. We observe that PA-MCTS, given appropriate $\alpha$, achieves significantly higher reward than both standard MCTS and DDQN under non-stationary settings. Also, as we observe in the setting where gravity is changed, PA-MCTS converges significantly faster than standard MCTS by leveraging the pre-trained policy. Finally, we observe that all approaches (PA-MCTS, standard MCTS, and DDQN) achieve close to 0 utility when the mass of the pole is set to 1.5kg, presumably because it is extremely difficult to balance the pole in this setting.


\subsection{Choosing the Optimal \texorpdfstring{$\alpha$}{Lg}}
We propose selecting $\alpha$ as before, i.e., we observe how PA-MCTS performs under different computational constraints set by the number of MCTS iterations. Again, we observe that the optimal $\alpha$ for 100 MCTS iterations is well approximated (almost perfectly) by 25 MCTS iterations. We show the results in \cref{{fig:masspole_pamcts}}.

\begin{figure}[!htb]
    \centering
    \includegraphics[width=0.6\textwidth, scale=1.0]{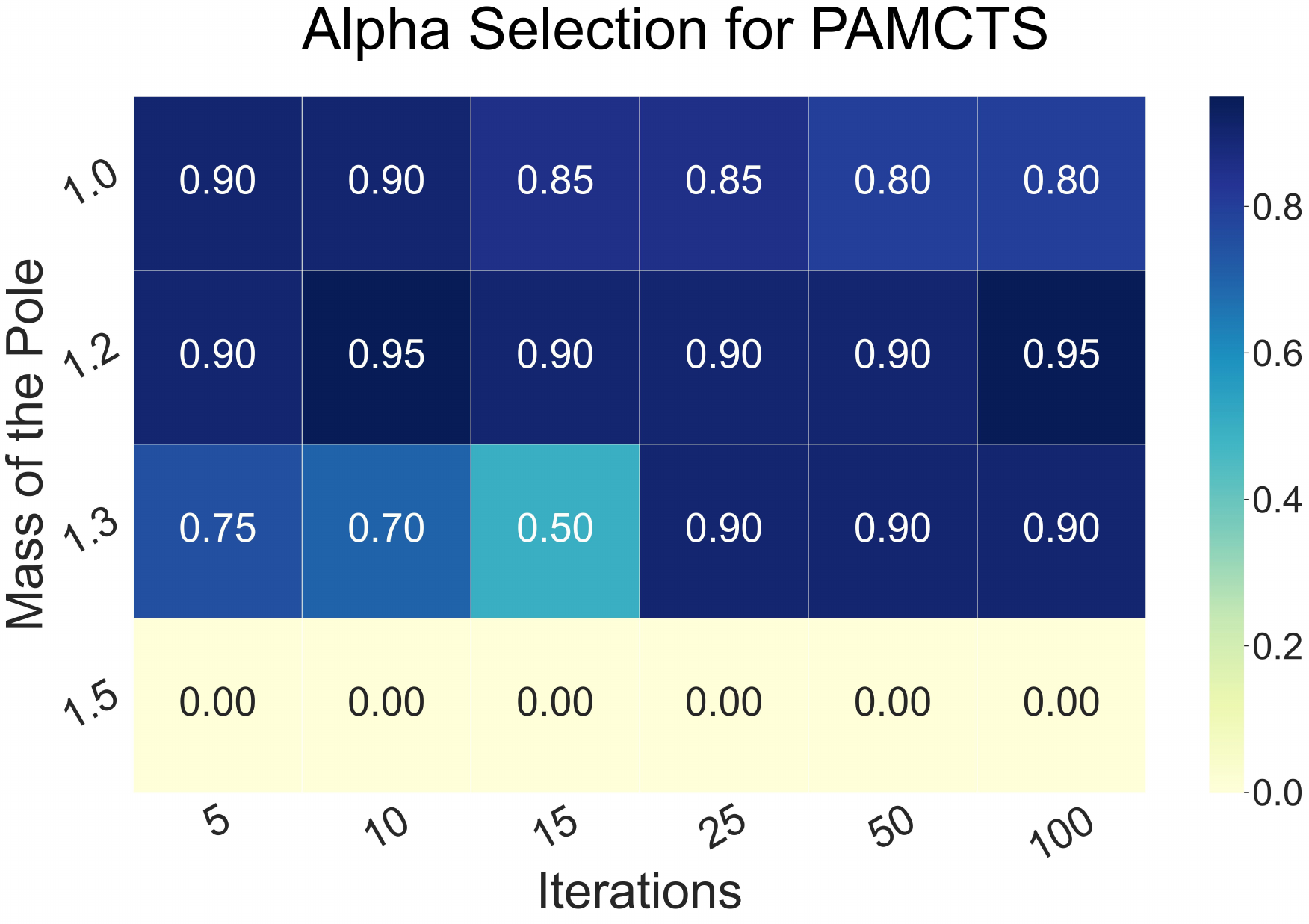}
    \caption{Cartpole Alpha Selection Changing Mass of the Pole among {1.0kg, 1.2kg, 1.3kg, 1.5kg}}.
    \label{fig:masspole_pamcts}
    \vspace{-0.5cm}
\end{figure}

\subsection{Comparison with Alphazero}

\begin{figure}[!htb]
	\centering
	\begin{subfigure}{.5\textwidth}
		\centering
		\includegraphics[width=\linewidth]{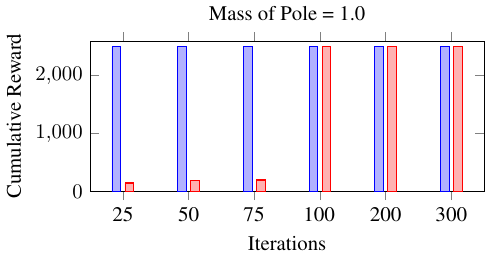}
	\end{subfigure}%
	\begin{subfigure}{.47\textwidth}
		\centering
		\includegraphics[width=\linewidth]{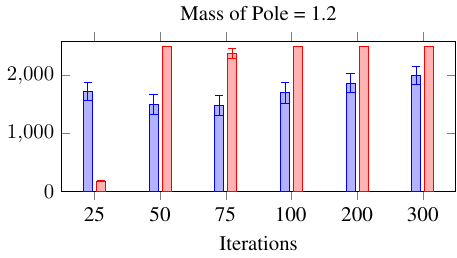}
	\end{subfigure}
	\begin{subfigure}{\textwidth}
		\centering
		\includegraphics[width=0.25\linewidth]{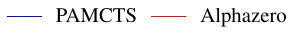}
		\caption*{} 
	\end{subfigure}
	
	\caption{We compare between the results of PA-MCTS with optimal alpha and the results of Alphazero}
	\label{figure:comparison_partial}
\end{figure}

We use the proposed $\alpha$-selection strategy to find the optimal alpha value for a given non-stationary setting, i.e., for each mass of the pole. Then, we compare PA-MCTS with the optimal $\alpha$ with Alphazero and present results in \cref{figure:comparison_partial}. Both AlphaZero and PA-MCTS achieve 0 utility when mass=1.5kg. When mass=1.0kg, we see that PA-MCTS outperforms AlphaZero by converging significantly faster. When mass=1.2kg, we see that AlphaZero slightly outperforms PA-MCTS.




\begin{figure}[!htb]
    \centering
    \includegraphics[width=\linewidth]{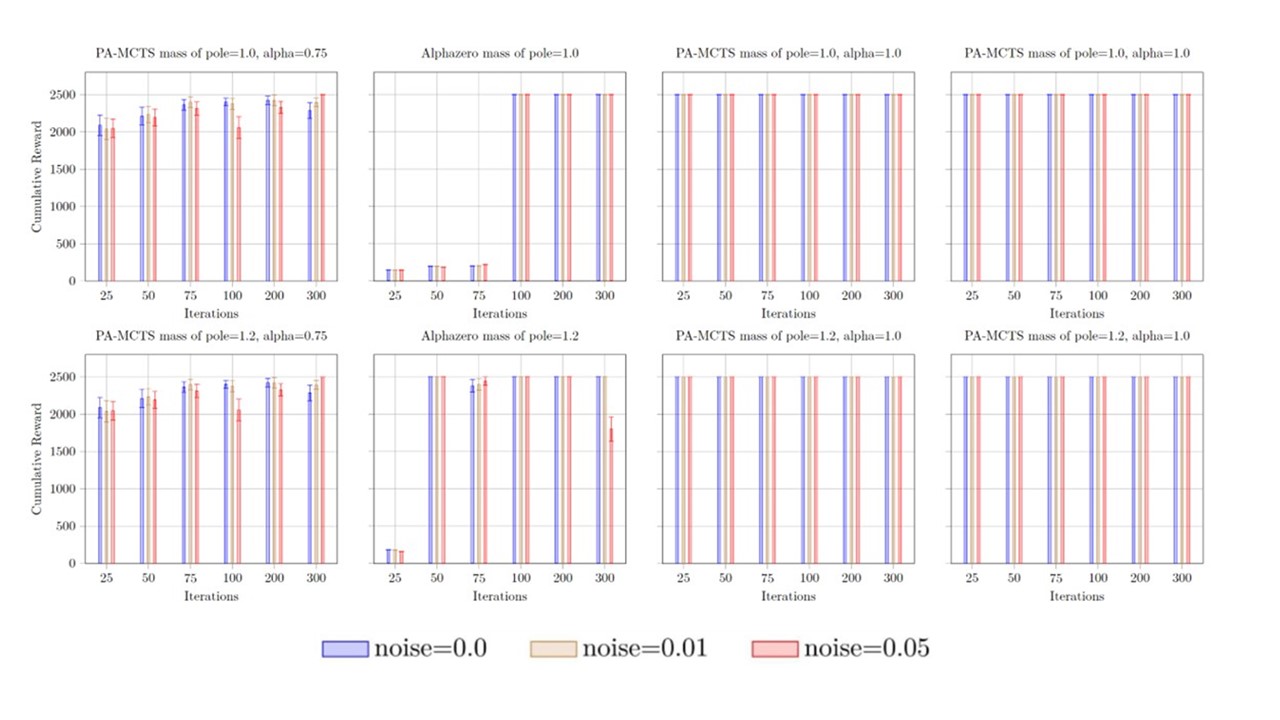}
    \caption{The performance of PA-MCTS and AlphaZero under varying levels of noise with mass of pole = 1.0kg, 1.2kg. PA-MCTS converges faster than Alphazero with small MCTS iterations.}
    \label{figure:masspole_bar_plots}
\end{figure}



\subsection{Noisy settings}
We now consider the setting that in practice, it is difficult to observe the environmental change exactly. As a result, we add noise between the environmental change and how the agent perceives the change. We introduce Gaussian noise (with 0 mean and varying standard deviation) in the mass, i.e., the agent perceives a noisy mass during simulation but the true non-stationary mass is used for execution. We show the results in \cref{figure:masspole_bar_plots}. We observe that even under noisy settings, PA-MCTS achieves higher rewards than AlphaZero in most cases, especially when the agent can afford low computational time for making decisions.


\subsection{Hardware}
The experiments were conducted on Chameleon testbed \citep{keahey2020lessons} on 4 Linux systems with 32-96 logical processors and 528 GB RAM.

\subsection{Extensive Results for Noisy Settings Changing Gravity}
\begin{figure}[!htb]
    \centering
    \includegraphics[width=\textwidth, scale=1.5]{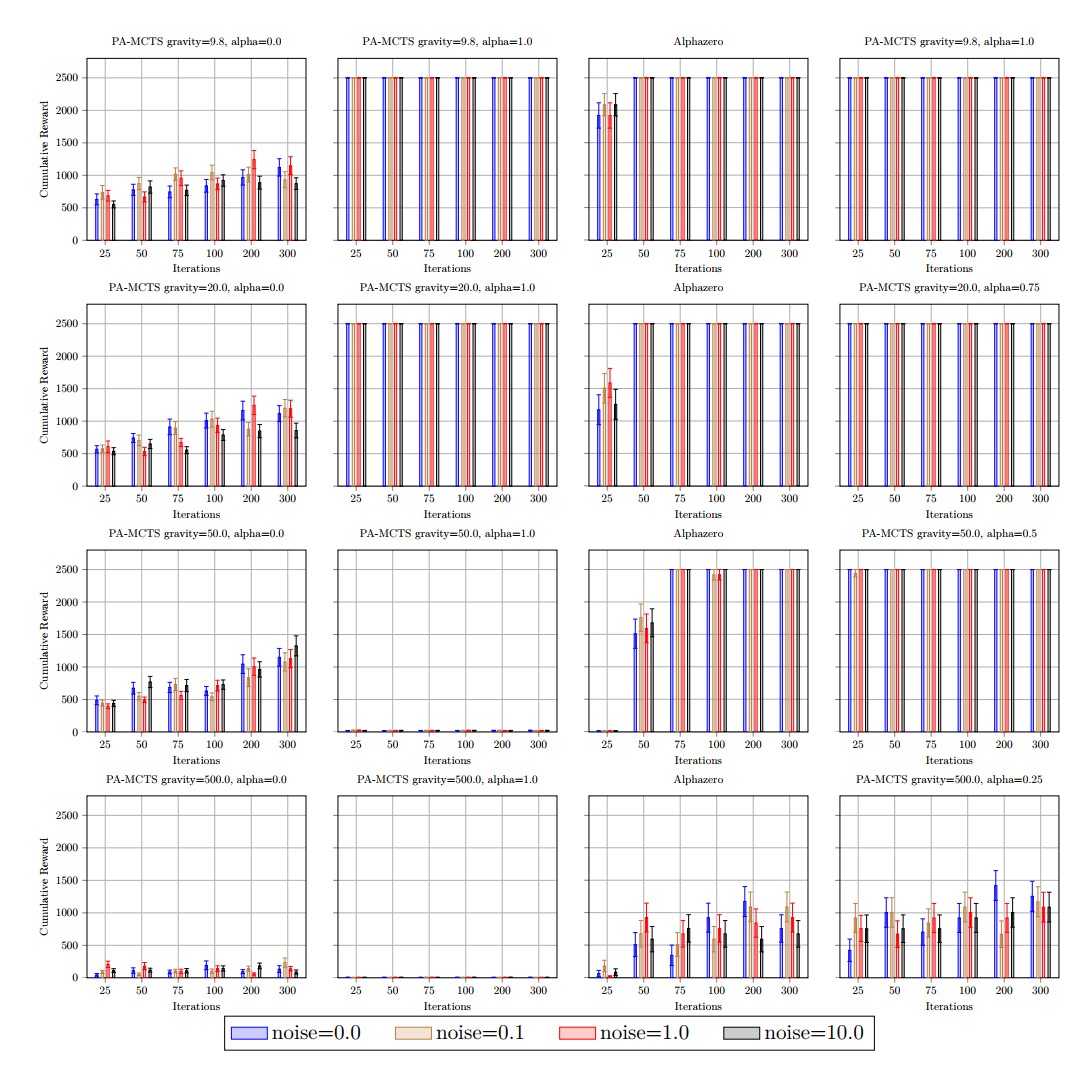}
    \caption{We compare the performance of PA-MCTS, Alphazero, Standard MCTS, DDQN for noise among 0.0, 0.1, 1, 10 with different gravity values among 9.8 $m/s^2$, 20.0 $m/s^2$, 50.0 $m/s^2$, 500.0 $m/s^2$}.
    \label{fig:full_gravity_bar_plots}
    \vspace{-0.5cm}
\end{figure}

In the main text, due to the space limitation, we only showed partial results for the noisy settings where gravity is changed. Here, we present complete results (for each value of updated gravity) with noisy settings. We show the extensive comparison results between PA-MCTS, Alphazero, Standard MCTS, and DDQN (as shown in \cref{fig:full_gravity_bar_plots}). We observe that (as in the main text), PA-MCTS outperforms the baselines even under noisy settings.

%% file: appendix_frozenlake.tex
\section{Experimental Results on the FrozenLake Environment} \label{app:frozenlake}


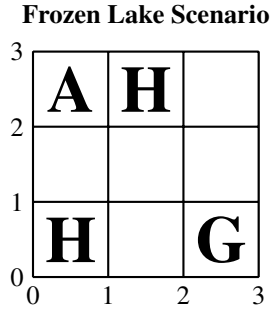
\begin{figure}[ht]
\centering
\begin{tikzpicture}
\centering
  \draw (0,0) grid (3,3); 
  \filldraw[fill=white!30] (0,0) rectangle (3,3);
  \foreach \x/\y/\label in {0/2/A,2/0/G, 1/2/H, 0/0/H}
    \node at (\x+0.5, \y+0.5) {\Huge\textbf{\label}};
  \foreach \x in {0,1,2,3}
    \node[anchor=north] at (\x,0) {\x};
  \foreach \y in {0,1,2,3}
    \node[anchor=east] at (0,\y) {\y};
  \node[align=center,font=\bfseries] at (1.5,3.5) {Frozen Lake Scenario};
  \draw[step=1, black, thick] (0,0) grid (3,3);
\end{tikzpicture}
\caption{The test environment for the FrozenLake experiments. The agent must traverse from cell (0,2) to cell (2,0) without falling into cell (0,0) and cell (1,2). However, under slippery conditions, the agent does not always move in the desired direction.}
\label{fig:frozen_lake_test_env}
\end{figure}

We also conduct experiments using the Frozenlake Environment. The goal of the agent is to travel on a slippery surface (a frozen lake) from a pre-defined starting position to a pre-defined goal position without falling into any holes. Due to the slippery nature of the surface, the agent does not always move in the intended direction.
We introduce non-stationarity by changing the probability of moving toward the desired direction. We vary this probability and set the probability of moving in the two perpendicular directions as equal. We represent an environment by a tuple of three probabilities which shows the \textbf{[probability to move towards the desired direction, probability to move towards the first perpendicular direction, probability to move towards the second perpendicular direction]}, e.g., $[1,0,0]$ refers to the stationary setting, where the surface is not slippery and the agent moves in its desired direction with certainty (i.e., probability of 1) and there are no chances of the agent moving in directions perpendicular to its desired direction (denoted by the two 0 entries). 
Given this setting, note that we can calculate $\eta$ (the amount of change in the probability transition function as defined in the main text); $\eta_{min} = 0$, when there is no change, i.e., the updated environment is the same as the stationary environment ($[1,0,0]$), and $\eta_{max} = 1.333$ when the updated environment is denoted by [0.333, 0.333, 0.333]. We show the environments we use for testing in \cref{prob_distribution_eta}. We choose non-stationary environments by uniformly sampling $\eta$ value from $\eta_{min}$ to $\eta_{max}$. We use a 3x3 map for Frozen Lake, with position indices from 0 to 8 as shown in the main text. The agent starts from cell 0, the goal is set at cell 8, and there are holes in cells 1 and 6. We show a schematic of the environment in \cref{fig:frozen_lake_test_env}.


\begin{table}[h]
\begin{center}
\begin{tabular}{@{}ll@{}}
\toprule
probability distribution  & $\eta$ \\ \midrule
{[}1.0, 0.0, 0.0{]}       & 0                   \\
{[}0.933, 0.033, 0.033{]} & 0.2                 \\
{[}0.833, 0.083, 0.083{]} & 0.4                 \\
{[}0.733, 0.133, 0.133{]} & 0.6                 \\
{[}0.633, 0.183, 0.183{]} & 0.8                 \\ 
{[}0.533, 0.233, 0.233{]} & 1.0                \\
{[}0.433, 0.283, 0.283{]} & 1.2                \\
{[}0.333, 0.333, 0.333{]} & 1.333               \\ \bottomrule
\end{tabular}
\end{center}
\caption{Probability distribution of the testing environments and corresponding $\eta$}
\label{prob_distribution_eta}
\end{table}

As \cref{prob_distribution_eta} shows, we choose probability distributions based upon uniformly sampling $\eta$ value from $\eta_{min}$ to $\eta_{max}$. 
The map used for this experiment is a 3x3 map, with position index from 0 to 8 and 0 is the start position, 8 is the destination, and position index 1 and 6 are holes. 


\begin{figure}[!htb]
	\centering
	\begin{subfigure}{.75\textwidth}
		\centering
		\includegraphics[width=\linewidth]{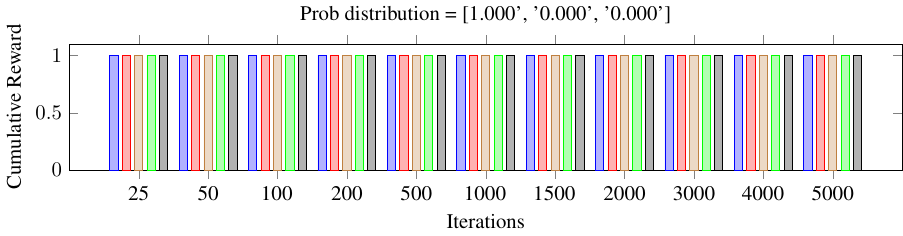}
	\end{subfigure} \\
	\begin{subfigure}{.75\textwidth}
		\centering
		\includegraphics[width=\linewidth]{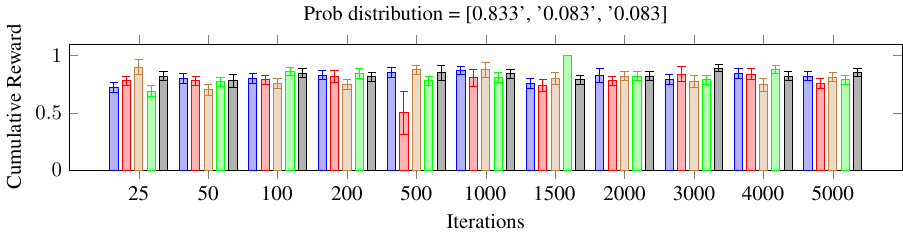}
	\end{subfigure} \\
	\begin{subfigure}{.75\textwidth}
		\centering
		\includegraphics[width=\linewidth]{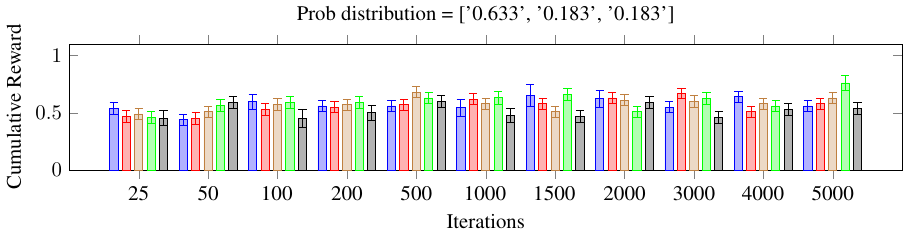}
	\end{subfigure}\\
	\begin{subfigure}{.75\textwidth}
		\centering
		\includegraphics[width=\linewidth]{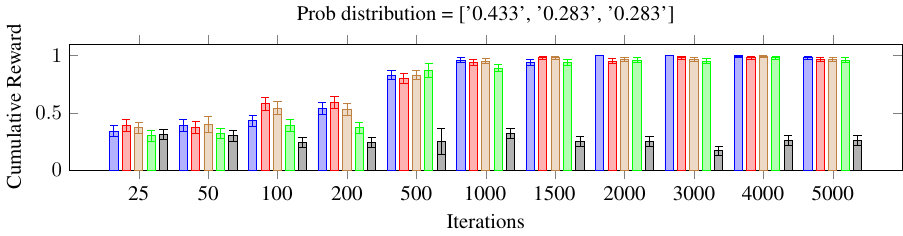}
	\end{subfigure}\\
	\begin{subfigure}{.75\textwidth}
		\centering
		\includegraphics[width=\linewidth]{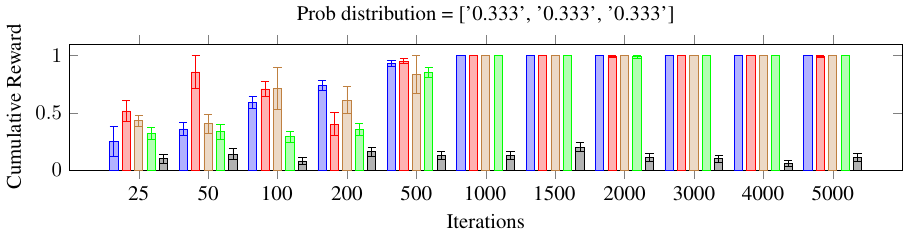}
	\end{subfigure}\\
	\begin{subfigure}{\textwidth}
		\centering
		\includegraphics[width=0.8\linewidth]{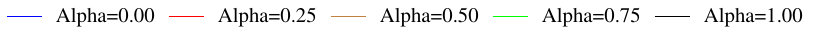}
		\caption*{} 
	\end{subfigure}
	
	\caption{We show the cumulative reward obtained by PA-MCTS, standard MCTS ($\alpha=0)$, and DDQN ($\alpha=1$ for different environmental changes.}
	\label{fig:frozenlake_pamcts}
\end{figure}

\subsection{Stationary Environment} \label{fl_stationary}
As described above, the stationary environment corresponds to the setting $[1,0,0]$, i.e., the lake is not slippery. As we see in \cref{fig:frozenlake_pamcts} shows, standard MCTS, PA-MCTS (with all values of $\alpha$) and DDQN can achieve maximum rewards. We compute each result by averaging across 100 samples.

\subsection{Non-Stationary Environment} \label{fl_nonstationary}
We also show results in non-stationary settings results in \cref{fig:frozenlake_pamcts}. We use different colors to represent different environments. We show results for five settings in the figure to make the plot readable (the results are consistent across the other settings). Like the cartpole environment, we observe that the performance of the DDQN deteriorates as the environment changes.

This behavior is expected, as actions that are favorable in the non-slippery setting can now lead the agent to a hole. We also see that for most settings, PA-MCTS achieves the optimal utility. In some settings, while PA-MCTS does not achieve the highest possible utility, it outperforms both standard MCTS and DDQN given the optimal $\alpha$ (we discuss how to tune $\alpha$ below).


\subsection{Choosing the optimal \texorpdfstring{$\alpha$}{Lg}}
\begin{figure}[!htb]
    \centering
    \includegraphics[width=0.6\textwidth, scale=1.2]{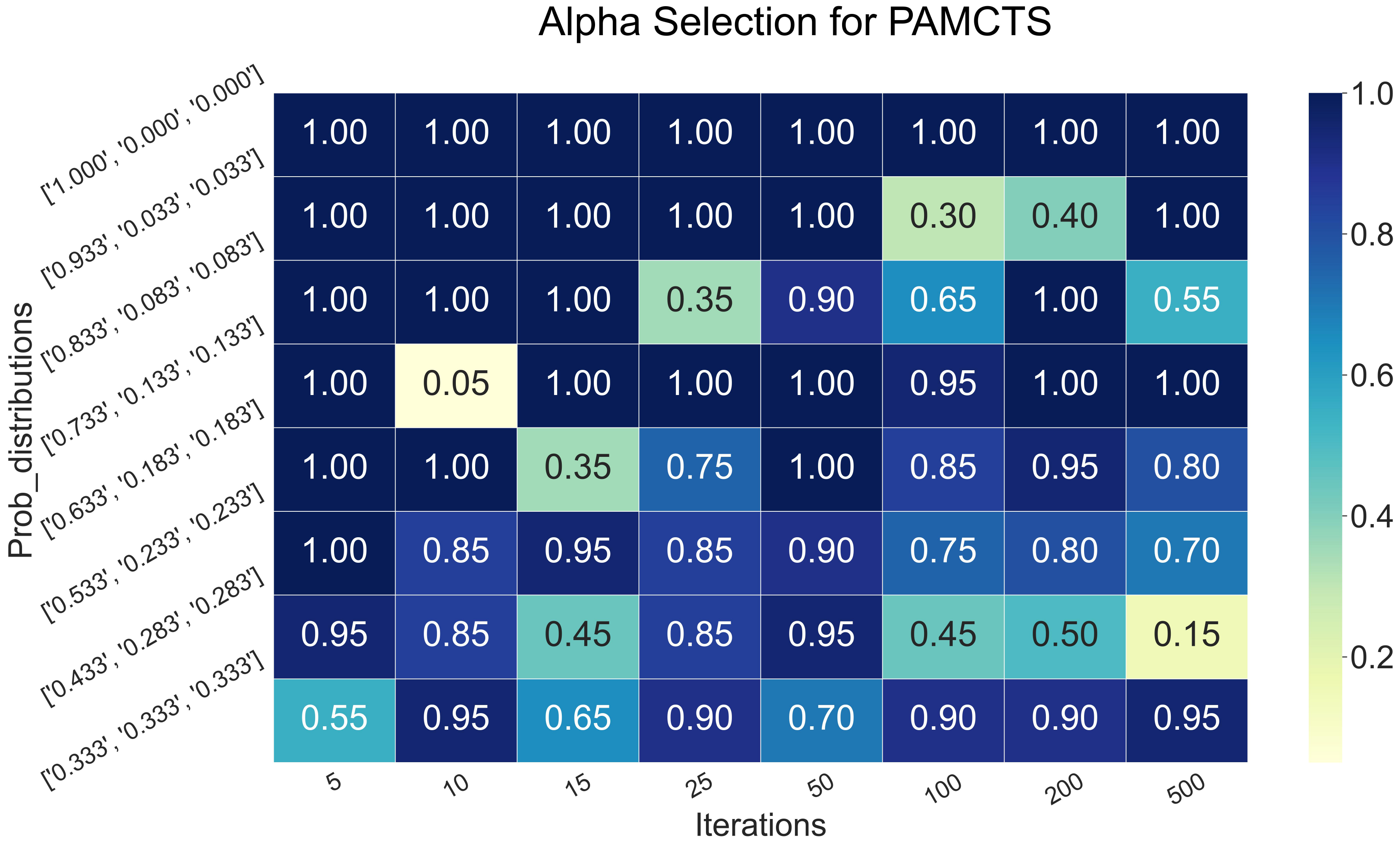}
    \caption{Frozenlake Alpha Selection for different probability distributions}.
    \label{fig:frozenlake_alpha_selection}
    \vspace{-0.5cm}
\end{figure}

As we describe in the main text, during execution, the agent needs to pick an optimal $\alpha$ quickly. Similar to the cartpole environment, we leverage the finding that the agent can approximate the performance of different values of $\alpha$ by using a small number of PA-MCTS iterations. We show the results in \cref{fig:frozenlake_alpha_selection}; note that PA-MCTS with 25 iterations approximates the optimal $\alpha$ with 500 iterations (naturally, this approach is not optimal; we hypothesize that the $\alpha$-selection mechanism can be operated in parallel even after an initial $\alpha$ is chosen, and the agent can update the parameters over time).

\subsection{Comparison with Alphazero}


\begin{figure}[!htb]
	\centering
	\begin{subfigure}{.75\textwidth}
		\centering
		\includegraphics[width=\linewidth]{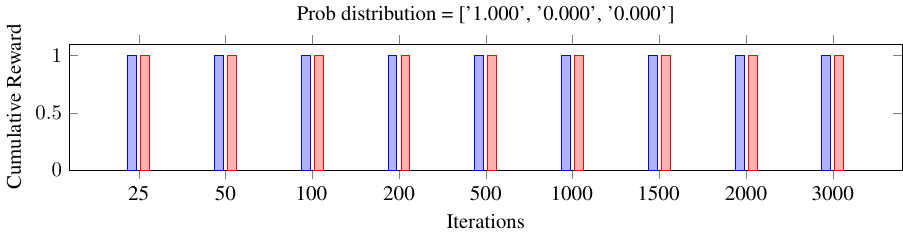}
	\end{subfigure}\\%
	\begin{subfigure}{.75\textwidth}
		\centering
		\includegraphics[width=\linewidth]{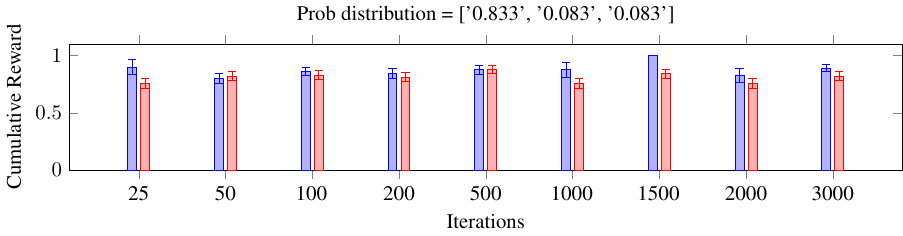}
	\end{subfigure}\\
	\begin{subfigure}{.75\textwidth}
		\centering
		\includegraphics[width=\linewidth]{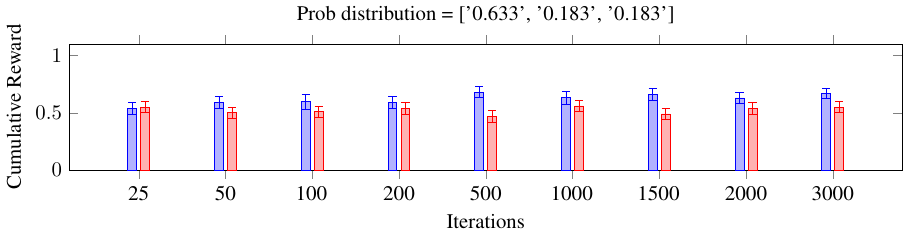}
	\end{subfigure}\\%
	\begin{subfigure}{.75\textwidth}
		\centering
		\includegraphics[width=\linewidth]{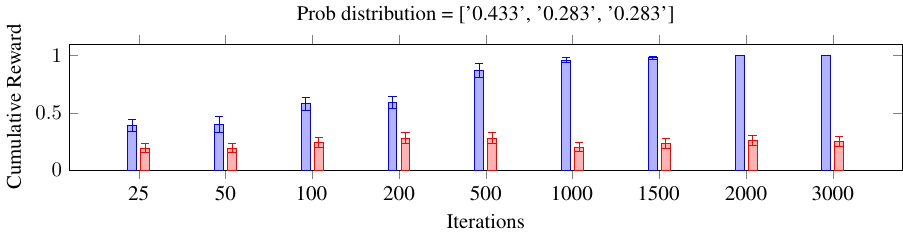}
	\end{subfigure}\\
	\begin{subfigure}{.75\textwidth}
		\centering
		\includegraphics[width=\linewidth]{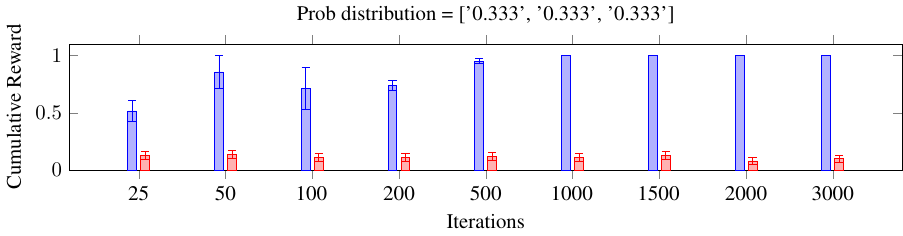}
	\end{subfigure}
	\begin{subfigure}{\textwidth}
		\centering
		\includegraphics[width=0.25\linewidth]{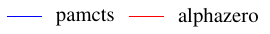}
		\caption*{} 
	\end{subfigure}
	
	\caption{Comparison of PA-MCTS with Alphazero; we see that PA-MCTS comprehensively outperforms Alphazero in non-stationary settings.}
	\label{fig:frozenlake_comparison}
\end{figure}

Finally, we use the proposed $\alpha$-selection strategy to select the optimal $\alpha$ value for different environment settings. Then, we compare PA-MCTS with Alphazero. \cref{fig:frozenlake_comparison} shows that as the non-stationary environment deviates further and further from the stationary environment, Alphazero performs significantly worse than PA-MCTS. We hypothesize this loss of performance is due to the fact that AlphaZero uses a pre-trained DDQN (on the stationary environment) to simulate rollouts, whereas PA-MCTS uses \textit{actual} rollouts as it grows the search tree.

\subsection{Time Constraints}
Finally, we explore the time taken to run PA-MCTS (with $\alpha$-selection, i.e., the additional time to tune $\alpha$ is included in the computational time we report) in comparison with the baseline approaches. Again, we point out that for a fair comparison, we re-train both DDQN and AlphaZero starting from the network trained on the stationary setting; we observed that this method significantly reduces the training time than re-training from scratch. We present the results in \cref{frozenlake_time_constraints}. Also, to be fair to the AlphaZero framework, we create 100 data generating processes in parallel to feed into its re-training procedure (we leverage Apache Kafka to achieve this framework). For the DDQN, we use a maximum of 1,000,000 million steps or stop the training if the network converges. For PA-MCTS, we run $\alpha$-selection in parallel. As we see, it is significantly faster to run PA-MCTS than re-training the networks.
\begin{table}[ht]
\begin{center}
\begin{tabular}{@{}ccc@{}}
\toprule
Method             & Iterations      & Execution Time (in seconds) \\ \midrule
PA-MCTS               & 5               & 0.001                       \\
PA-MCTS               & 10              & 0.002                       \\
PA-MCTS               & 15              & 0.005                       \\
PA-MCTS               & 25              & 0.006                       \\
PA-MCTS               & 50              & 0.007                       \\
PA-MCTS               & 100             & 0.012                       \\
PA-MCTS               & 200             & 0.029                       \\
PA-MCTS               & 500             & 0.08                        \\
PA-MCTS               & 1000            & 0.3                         \\
PA-MCTS               & 1500            & 0.41                        \\
PA-MCTS               & 2000            & 0.645                       \\
PA-MCTS               & 3000            & 0.856                       \\
Re-Train DDQN       & 1,000,000 steps & 1865.75                     \\
Re-Train AlphaZero & NA              & 3900                        \\ \bottomrule
\end{tabular}
\caption{Execution time of PA-MCTS (with $\alpha$-selection) with varying MCTS iterations in comparison with DDQN and AlphaZero. We observe that PA-MCTS is significantly faster.}
\label{frozenlake_time_constraints}
\end{center}
\end{table}

\subsection{Hardware} \label{sec:fl_hardware}
The experiments were conducted on Chameleon testbed \citep{keahey2020lessons} on 4 Linux systems with 32-96 logical processors and 528 GB RAM.

%% file: appendix_cliffwalking.tex
\section{Experimental Results on the Cliff Walking Environment} 
\label{app:cliffwalking}
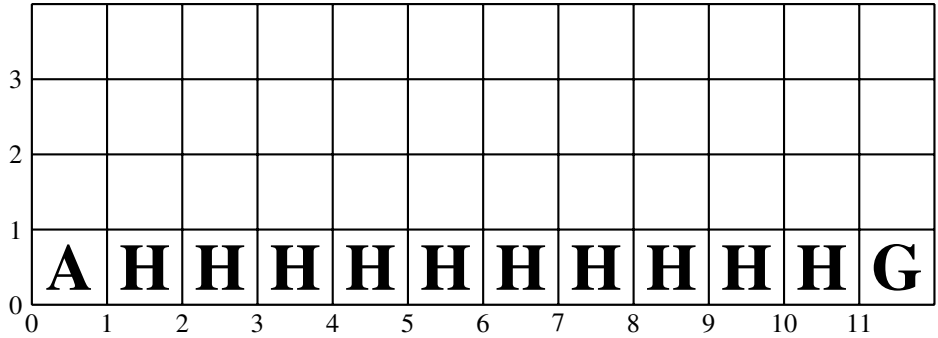
\begin{figure}[!htb]
\centering
\begin{tikzpicture}
\centering
  \draw (0,0) grid (12,4); 
  \filldraw[fill=white!30] (0,0) rectangle (12,4);
  \foreach \x/\y/\label in {0/0/A,11/0/G, 1/0/H, 2/0/H, 3/0/H, 4/0/H, 5/0/H, 6/0/H, 
  7/0/H, 8/0/H, 9/0/H, 10/0/H}
    \node at (\x+0.5, \y+0.5) {\Huge\textbf{\label}};
  \foreach \x in {0,1,2,3, 4, 5, 6, 7, 8, 9, 10, 11}
    \node[anchor=north] at (\x,0) {\x};
  \foreach \y in {0,1,2,3}
    \node[anchor=east] at (0,\y) {\y};
  \node[align=center,font=\bfseries] at (5.5,4.5) {Cliff Walking Scenario};
  \draw[step=1, black, thick] (0,0) grid (12,4);
\end{tikzpicture}
\caption{The test environment for the Cliff Walking experiments. The agent must traverse from cell (1,0) to cell (10,0) without falling into the cliff. Under Slippery conditions, the agent does not always go in the desired direction.}
    \label{fig:cliff_walking_test_env}
\end{figure}

We also conduct experiments on the Cliff Walking Environment. The environment has 48 possible states. Instead of Cliff Walking implemented by OpenAI Gymnasium~\cite{brockman2016openai}, which only has deterministic transition functions, we implement our own Cliff Walking environment, which has stochastic transition functions (our code is available through the supplementary material). We show a schematic of the environment in \cref{fig:cliff_walking_test_env}; cell 36 is the start position, and state 47 is defined as the goal position. States in {37, 38, 39, 40, 41, 42, 43, 44, 45, 46}, shown in brown color in \cref{fig:cliff_walking_test_env} are defined as cliff region. Unlike OpenAI Gymnasium version~\cite{brockman2016openai}, our environment ends when the agent falls off to a cliff, or the agent reaches the goal position. The agent receives a reward only if it reaches the goal position, and the corresponding reward is related to how many steps the agent takes to reach the goal position from the start position. The fewer steps the agent takes, the more reward is received. For instance, the agent receives more reward by choosing path [36, 24, 25, 26, 27, 28, 29, 30, 31, 32, 33, 34, 35, 47] than choosing path [36, 24, 12, 0, 1, 2, 3, 4, 5, 6, 7, 8, 9, 10, 11, 23, 35, 47]. We introduce a slippery nature in this environment in the same manner as the Frozen Lake environment, with one critical difference to account for the fact that the hill has a specific location. If the slippery factor is 0.1, then there is a 0.1 probability that the agent goes to the cell below its current location (except when it explicitly wants to go down, in which case it always goes down). 


\begin{figure}[!htb]
	\centering
	\begin{subfigure}{.5\textwidth}
		\centering
		\includegraphics[width=\linewidth]{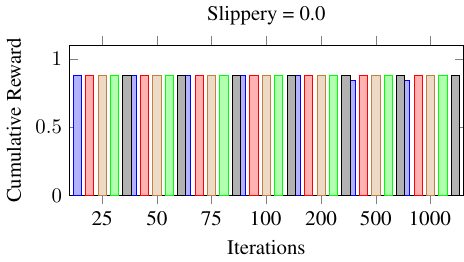}
	\end{subfigure}%
	\begin{subfigure}{.47\textwidth}
		\centering
		\includegraphics[width=\linewidth]{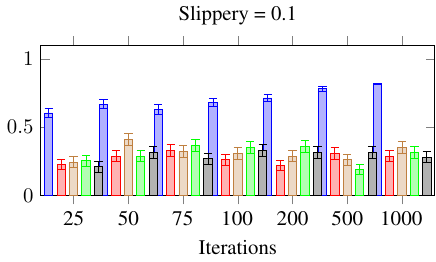}
	\end{subfigure}
	\begin{subfigure}{.5\textwidth}
		\centering
		\includegraphics[width=\linewidth]{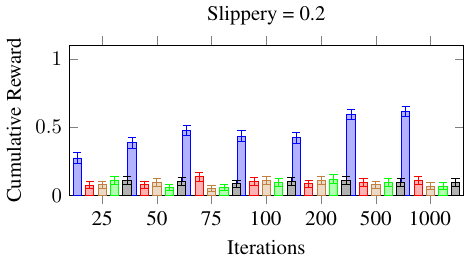}
	\end{subfigure}%
	\begin{subfigure}{.47\textwidth}
		\centering
		\includegraphics[width=\linewidth]{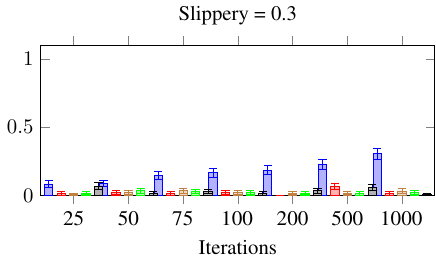}
	\end{subfigure}
	\begin{subfigure}{\textwidth}
		\centering
		\includegraphics[width=0.8\linewidth]{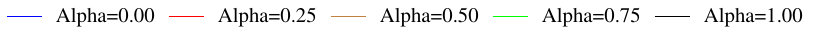}
		\caption*{} 
	\end{subfigure}
	
	\caption{We show the cumulative reward obtained by PA-MCTS, standard MCTS ($\alpha=0$), and DDQN ($\alpha=1$) for different slippery levels among 0.0, 0.1, 0.2, 0.3}
	\label{fig:cliff_walking_pamcts}
\end{figure}

\subsection{Stationary Environment} \label{cliff_walking_stationary}
The stationary environment is defined as the slippery factor is 0, i.e. the environment is not slippery. As we see in \cref{fig:cliff_walking_pamcts}, both PA-MCTS and DDQN can achieve maximum rewards. There is some noise in the performance of standard MCTS. We believe this randomness could be alleviated by introducing more iterations.We compute each result by averaging across 100 samples.

\subsection{Non-Stationary Environment} \label{cliff_walking_nonstationary_1}
In \cref{fig:cliff_walking_pamcts}, we also show results in non-stationary settings. We use different colors to represent different $\alpha$ values used for PA-MCTS. Similar to the Frozen Lake environment, the performance of the DDQN deteriorates as the environment changes. This behavior is expected, as actions that are favorable in the non-slippery setting can now lead the agent to the cliff region. As the surface becomes more slippery, we observe that using standard MCTS is better than leveraging learned Q-values. Indeed, PA-MCTS selects 0 as the optimal $\alpha$, disregarding the learned Q-values. This shows the robustness of PA-MCTS, i.e., when the learned Q-values are not useful, it reverts to a fully online search.

\subsection{Choosing the optimal \texorpdfstring{$\alpha$}{Lg}}
\begin{figure}[!htb]
    \centering
    \includegraphics[width=0.6\textwidth]{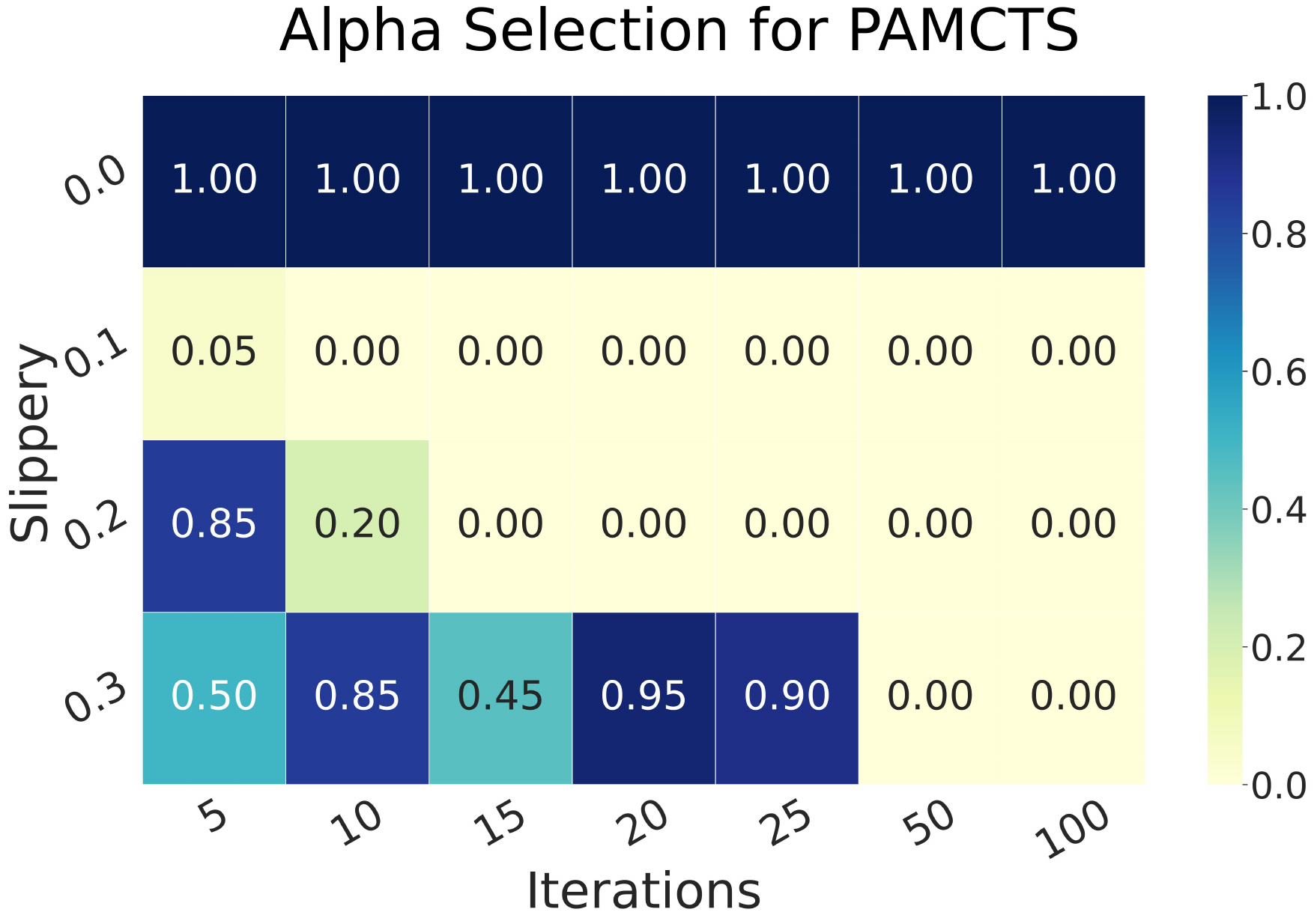}
    \caption{Cliff Walking Alpha Selection for different Slippery Levels}.
    \label{fig:cliff_walking_alpha_selection}
    \vspace{-0.5cm}
\end{figure}

During execution, the agent needs to pick an optimal $\alpha$ quickly. Similar to previous environments, we utilize the observation that the agent can emulate the performance outcomes from various $\alpha$ values by executing a limited count of PA-MCTS iterations. The outcomes are displayed in \cref{fig:cliff_walking_alpha_selection}. For a stationary environment, counting on DDQN could give a maximum cumulative reward. In non-stationary environments, the selection is more biased towards $alpha=0.0$, as the optimal path chosen by DDQN becomes risky. Given 50 iterations, MCTS results become more reliable.

\subsection{Comparison with Alphazero}


\begin{figure}[!htb]
	\centering
	\begin{subfigure}{.5\textwidth}
		\centering
		\includegraphics[width=\linewidth]{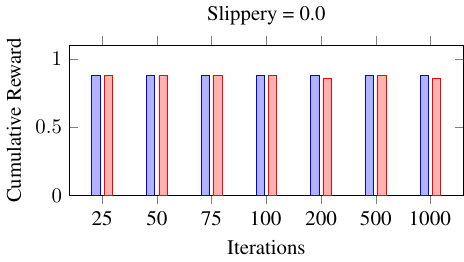}
	\end{subfigure}%
	\begin{subfigure}{.47\textwidth}
		\centering
		\includegraphics[width=\linewidth]{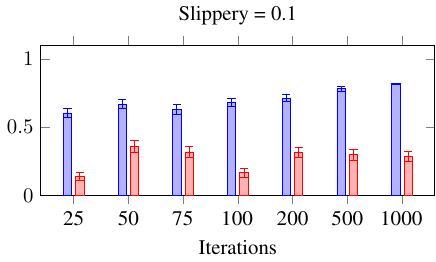}
	\end{subfigure}
	\begin{subfigure}{.5\textwidth}
		\centering
		\includegraphics[width=\linewidth]{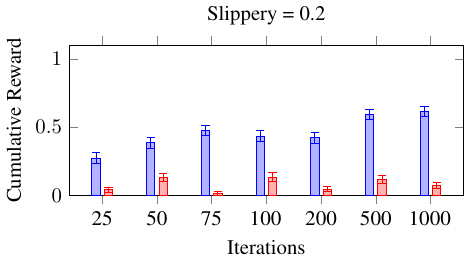}
	\end{subfigure}%
	\begin{subfigure}{.47\textwidth}
		\centering
		\includegraphics[width=\linewidth]{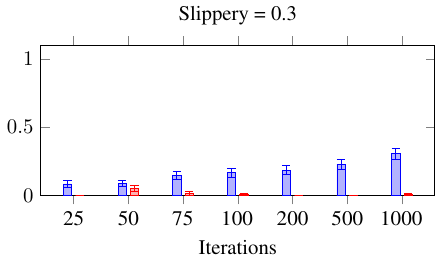}
	\end{subfigure}
	\begin{subfigure}{\textwidth}
		\centering
		\includegraphics[width=0.25\linewidth]{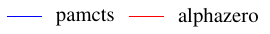}
		\caption*{} 
	\end{subfigure}
	
	\caption{Comparison of PA-MCTS with Alphazero; we see that PA-MCTS comprehensively outperforms Alphazero in non-stationary settings.}
	\label{fig:cliff_walking_comparison}
\end{figure}

We compare PA-MCTS with Alphazero after we use the proposed $\alpha$-selection strategy to find the optimal $\alpha$ value for different environment settings. In \cref{fig:cliff_walking_comparison}, as slippery becomes nonzero, alphazero performance becomes significantly worse than PA-MCTS. This is reasonable as Alphazero selects actions in UCT by selecting actions preferred by those in stationary settings. 

\subsection{Time Constraints}
As in previous environments, we explore the time taken to run PA-MCTS (with $\alpha$-selection) and compare it with the time to retrain DDQN and Alphazero. We show the results in \cref{cliff_walking_time_constraints}. We retrain DDQN and Alphazero that were previously trained in the stationary environment setting. For alphazero network, we create 60 data-generating processes in parallel to feed into its re-training procedure. For DDQN, we use a maximum of 100,000 steps to stop the training. As we did previously, we run $\alpha$-selection in parallel, and running PA-MCTS is significantly faster than retraining the networks.

\begin{table}[ht]
\begin{center}
\begin{tabular}{@{}ccc@{}}
\toprule
Method             & Iterations      & Execution Time (in seconds) \\ \midrule
PA-MCTS               & 5               &  0.019                \\
PA-MCTS               & 10              &  0.032                \\
PA-MCTS               & 15              &  0.038              \\
PA-MCTS               & 20              &  0.392              \\
PA-MCTS               & 25              &  0.401             \\
PA-MCTS               & 50              &  0.071                 \\
PA-MCTS               & 100             &  0.113              \\
PA-MCTS               & 200             &  0.218                  \\
PA-MCTS               & 500             &  0.570               \\
PA-MCTS               & 1000            &  0.919                      \\
Re-Train DDQN       & 100,000 steps &   2916.864                   \\
Re-Train AlphaZero & NA              &   6097.156                \\ \bottomrule
\end{tabular}
\caption{Execution time of PA-MCTS (with $\alpha$-selection) with varying MCTS iterations in comparison with DQN and AlphaZero. We observe that PA-MCTS is significantly faster.}
\label{cliff_walking_time_constraints}
\end{center}
\end{table}

\subsection{Hardware} \label{sec:cliff_walking_hardware_1}
The experiments were conducted on Chameleon testbed \citep{keahey2020lessons} on 4 Linux systems with 32-96 logical processors and 528 GB RAM.

%% file: appendix_lunarlander.tex
\section{Experimental Results on the Lunar Lander Environment} 
We also evaluated our approach on the Lunar Lander Environment from OpenAI Gymnasium~\cite{brockman2016openai}. In this setting, the agent must control the lunar module by firing the main engine upwards or the side engines to steer left or right. We use the discrete version of the environment. The state is an 8-dimensional vector: the coordinates of the lander, its linear velocities in its angle, its angular velocity, and two booleans that represent whether each leg is in contact with the ground or not. We change the windpower parameter in order to represent uncertainty in this environment.

\begin{figure}[!htb]
    \centering
    \includegraphics[width=0.5\textwidth]{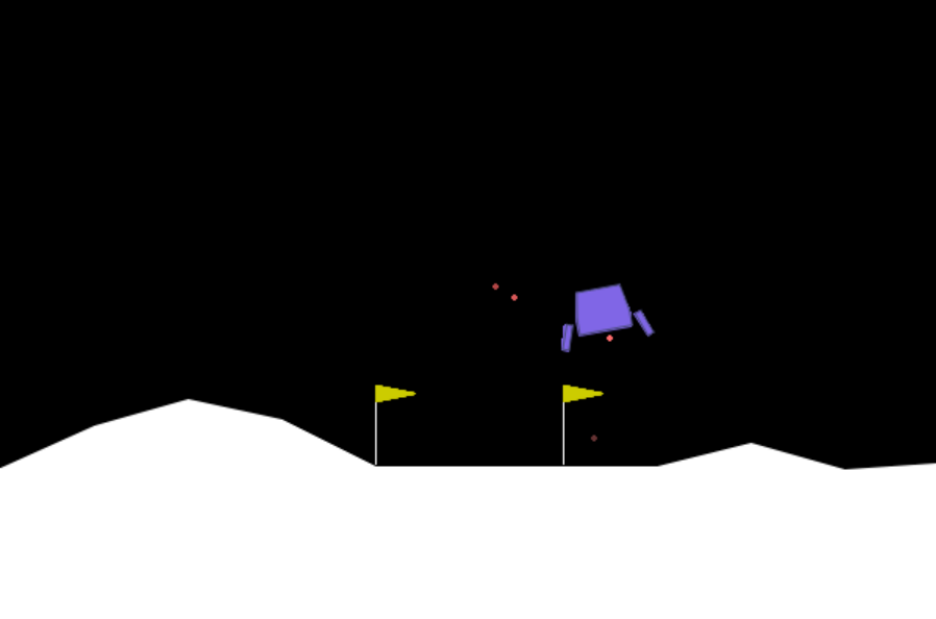}
    \caption{The test environment for the Lunar Lander Environment. The agent must land within a given viewpoint without crashing the body.}
    \label{fig:lunar_lander_test_env}
\end{figure}


\begin{figure}[!htb]
	\centering
	\begin{subfigure}{.5\textwidth}
		\centering
		\includegraphics[width=\linewidth]{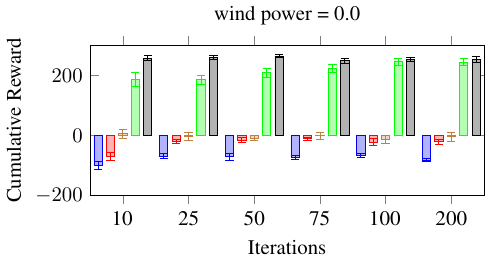}
	\end{subfigure}%
	\begin{subfigure}{.47\textwidth}
		\centering
		\includegraphics[width=\linewidth]{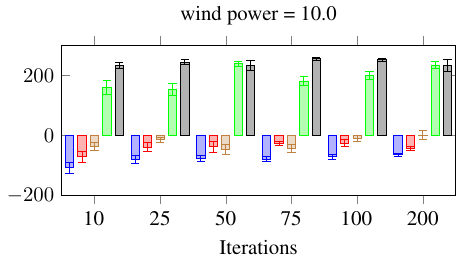}
	\end{subfigure}
	\begin{subfigure}{.5\textwidth}
		\centering
		\includegraphics[width=\linewidth]{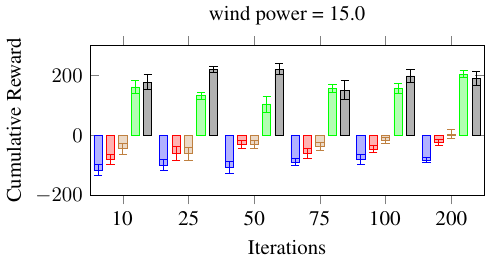}
	\end{subfigure}%
	\begin{subfigure}{.47\textwidth}
		\centering
		\includegraphics[width=\linewidth]{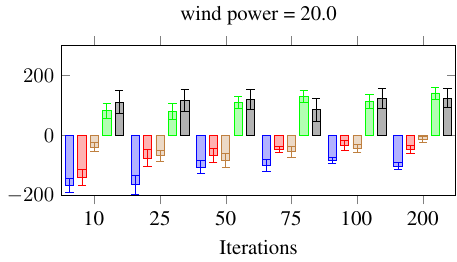}
	\end{subfigure}
	\begin{subfigure}{\textwidth}
		\centering
		\includegraphics[width=0.8\linewidth]{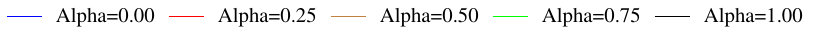}
		\caption*{} 
	\end{subfigure}
	
	\caption{We show the cumulative reward obtained by PA-MCTS, standard MCTS ($\alpha=0$), and DDQN ($\alpha=1$) for changing wind power among \{0.0, 10.0, 15.0, 20.0\}.}
	\label{fig:lunar_lander_pamcts}
\end{figure}

\subsection{Stationary Environment} \label{lunar_lander_stationary}
The stationary setting in this environment has no wind, i.e., the wind power is set to 0. As \cref{fig:lunar_lander_pamcts} shows, both DDQN and PA-MCTS with higher $\alpha$ value can achieve relatively high cumulative reward. We compute each result by averaging across 30 samples.

\subsection{Non-Stationary Environment} \label{cliff_walking_nonstationary}
We show results for different wind power values in \cref{fig:lunar_lander_pamcts} as well. We use different colors to represent different $\alpha$ values used for PA-MCTS. As the results suggest, it is challenging to use MCTS with a limited number of iterations to find near-optimal actions. This environment serves as the other end of the spectrum (with respect to cliff walking), where MCTS is not particularly useful given limited computational budget. In this case, relying on learned Q-values is the best course of action. Below, we show that PA-MCTS indeed selects a very high $\alpha$, therefore relying more on the learned Q-values.

\subsection{Choosing the optimal \texorpdfstring{$\alpha$}{Lg}}
As we expected, since the limited number of iterations can not fully explore the future trajectory of the agent, PA-MCTS relies more on the predictions from trained DDQN. 

\begin{figure}[!htb]
    \centering
    \includegraphics[width=0.5\textwidth]{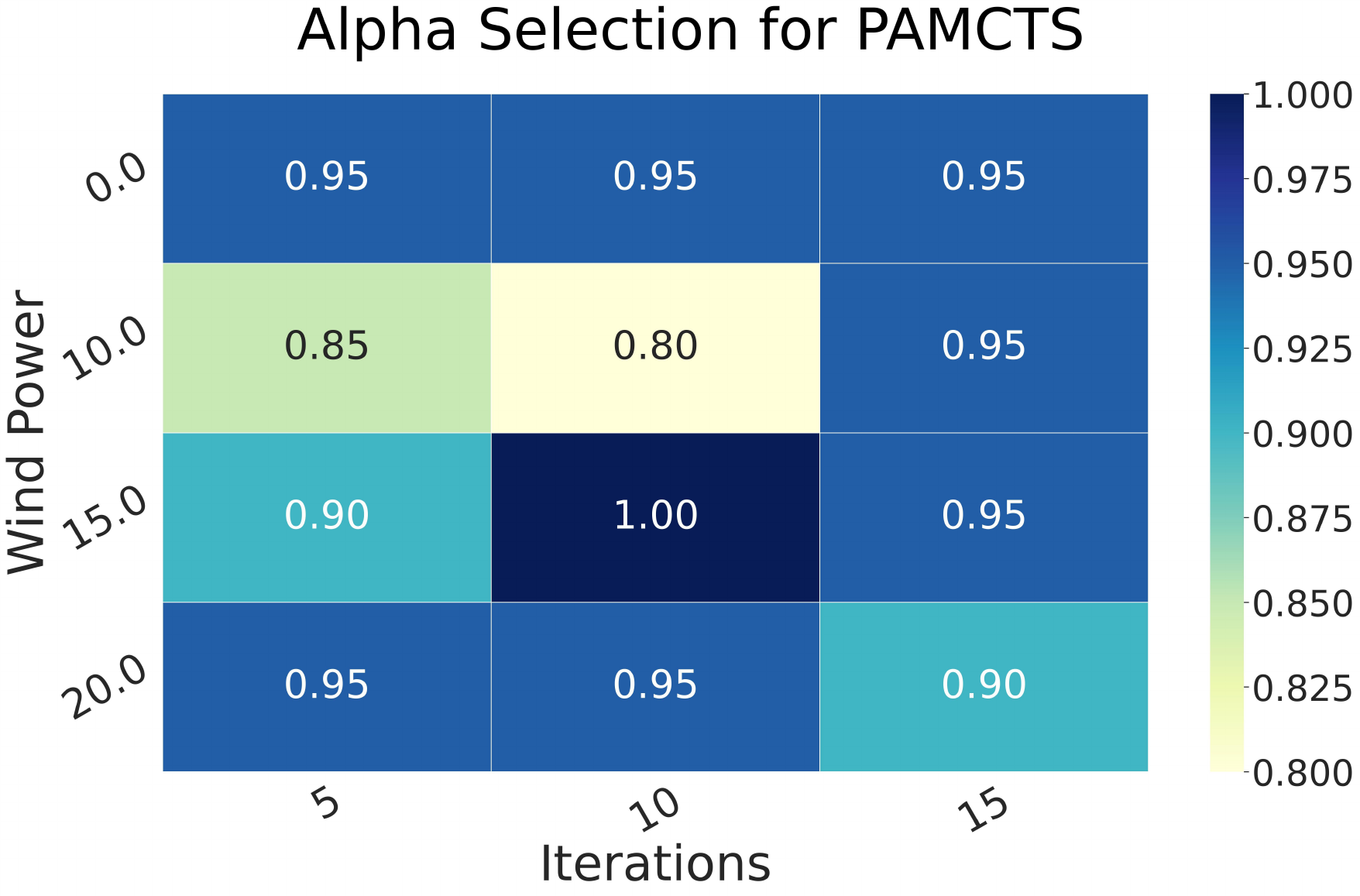}
    \caption{Lunar Lander Alpha Selection for different Wind Power}.
    \label{fig:lunar_lander_alpha_selection}
    \vspace{-0.5cm}
\end{figure}

\subsection{Comparison with Alphazero}
We deployed Alphazero training with 90 parallel running data-generating processes that feed data to the training process for 96 hours. However, in our setting, the AlphaZero network, even with multiple hyperparameters, did not converge and showed worse results than DDQN


\subsection{Hardware} \label{sec:cliff_walking_hardware}
The experiments were conducted on Chameleon testbed \citep{keahey2020lessons} on 4 Linux systems with 32-96 logical processors and 528 GB RAM.